\documentclass[draft]{agujournal2019}
\usepackage{url} 
\usepackage{lineno}
\usepackage[inline]{trackchanges} 
\usepackage{soul}
\usepackage{float}
\usepackage{epstopdf, epsfig}
\usepackage{caption}
\usepackage{subcaption}
\usepackage{amsmath}
\usepackage{amssymb}
\usepackage{color}
\usepackage{placeins}
\usepackage{dcolumn}%
\usepackage{bm}%
\usepackage{enumitem}

\draftfalse

\journalname{Journal of Advances in Modeling Earth Systems (JAMES)}
\begin{document}
\title{A probabilistic framework for learning non-intrusive corrections to long-time climate simulations from short-time training data}
%
%
\authors{B. Barthel Sorensen\affil{1}, L. Zepeda-Núñez\affil{2, 3},I. Lopez-Gomez\affil{2}, Z. Y. Wan\affil{2}, R. Carver\affil{2}, F. Sha\affil{2}, and T. P. Sapsis\affil{1}}

\affiliation{1}{Massachusetts Institute of Technology, Cambridge, MA 02139, USA}
\affiliation{2}{Google Research, 1600 Amphitheatre Parkway, Mountain View, CA 94043, USA}
\affiliation{3}{University of Wisconsin-Madison, 480 Lincoln Drive, Madison, WI 53706, USA}

\correspondingauthor{B. Barthel Sorensen}{bbarthel@mit.edu}



\begin{keypoints}
\item We present a probabilistic framework for debiasing coarse-resolution climate simulations using machine learning
\item The method accurately predicts the risk of events with return periods far longer than the training period
\item The method leverages probabilistic machine learning architectures to provide built-in uncertainty quantification
\end{keypoints}

\begin{abstract}
\textcolor{black}{
Despite advances in high performance computing, accurate numerical simulations of global atmospheric dynamics remain a challenge. The resolution required to fully resolve the vast range scales as well as the strong coupling with -- often not fully-understood -- physics renders such simulations computationally infeasible over time horizons relevant for long-term climate risk assessment. While data-driven parameterizations have shown some promise of alleviating these obstacles, the scarcity of high-quality training data and their lack of long-term stability typically hinders their ability to capture the risk of rare extreme events. In this work we present a general strategy for training variational (probabilistic) neural network (NN) models to non-intrusively correct under-resolved long-time simulations of turbulent climate systems. The approach is based on the paradigm introduced by \citeA{barthel_sorensen_non-intrusive_2024} which involves training a post-processing correction operator on under-resolved simulations nudged towards a high-fidelity reference. Our variational framework enables us to learn the dynamics of the underlying system from very little training data and thus drastically improve the extrapolation capabilities of the previous deterministic state-of-the art -- even when the statistics of that training data are far from converged. We investigate and compare three recently introduced variational network architectures and illustrate the benefits of our approach on an anisotropic quasi-geostrophic flow. For this prototype model our approach is abe to not only accurately capture global statistics, but also the anistropic regional variation and the statistics of multiple extreme event metrics -- demonstrating significant improvement over previously introduced deterministic architectures.
}


\end{abstract}

\section*{Plain Language Summary}
We present a probabilistic framework to build and train machine learned (ML) correction operators to improve the predicted statistics of low-resolution climate simulations. The proposed methodology is specifically focused on enabling long-time climate predictions using operators trained on short-time data. We illustrate our approach, which acts on existing data in a post-processing manner, on a prototype climate model, for which we are able to accurately quantify the regionally varying statistics as well as rare-event statistics \textcolor{black}{over the previous state-of-the-art}. The simple model we consider here allows us to demonstrate our method on very long simulation, but our method can be readily applied to output from full-complexity climate models.

\section{Introduction}
As the Earth's climate changes, we are faced with deep uncertainty about extreme weather events whose frequency and magnitude are expected to increase \cite{Lehmann2015,meehl2004,perkins_kirkpatrick_2020}. Due to their potential for catastrophic consequences, it is crucial to accurately quantify their long-term risk and assess their impact on communities \cite{fischer_increasing_2021,raymond_understanding_2020,robinson_increasing_2021}. 
In this context, ``extreme events'' are generally defined as high amplitude anomalies of high-impact variables, such as near-surface temperature and precipitation \cite{lucarini_extremes_2016,sapsis_statistics_2021}, to which human activities are highly sensitive. For instance, heatwaves
can have devastating effects on an unprepared population, particularly when compounded with other events such as low rainfall \cite{bevacqua_advancing_2023,raymond_understanding_2020,robinson_increasing_2021,zscheischler_future_2018}. 

From a statistical point of view, certain observables being susceptible to extreme events implies that their probability density functions (pdfs) have ``heavy tails'', i.e. they decay slowly and high amplitude events retain small but non-negligible probability.
Accurately quantifying the risk of such events is subject to two main requirements: first, high-fidelity simulations that can capture the dynamics of interest, which require a high-resolution mesh in space and time; and second, sufficiently large samples to capture rare events in the tail of the distribution. The latter can be obtained either through long-term simulations or through large simulation ensembles \cite{Deser2012}. However, due to the high-dimensional, chaotic, and multi-scale nature of Earth's atmosphere, large ensembles of high-resolution simulations are computationally intractable over multi-decade or multi-century time horizons. As an example, the highest resolution climate models currently proposed fall short of fully resolving all the spatial scales of atmospheric turbulence by a factor of $10^{17}$ degrees of freedom \cite{schneider_harnessing_2023}. These shortfalls are further compounded by the need to simulate centuries-long trajectories for climate risk assessment. 

Alternative surrogate machine learning (ML) techniques are becoming increasingly attractive as a computationally efficient way to simulate the Earth's atmosphere \cite{pathak2022fourcastnet}. Alas, purely data-driven models present their own set of challenges. In contrast to dynamical models, they are often unstable when run over long time-horizons, and they struggle to extrapolate beyond the distribution defined by the scarce training data. This becomes problematic as we seek to quantify climate risks over the coming centuries with only several decades of observational data available for training. Although methodologies have been proposed to circumvent these issues by exploiting properties of the underlying dynamical system \cite{kochkov2023neural,mathews2021uncovering}, most of them require an explicit notion of ergodicity \cite{li2022learning,jiang2023training,platt2023constraining,schiff2024:dyslim}, or scale poorly as the state dimension increases \cite{pathak2017using, bollt2021explaining, hara2022learning}, posing challenges for their use in climate-related applications.

These limitations have spurred a complementary line of research in which hybrid strategies are explored \cite{schneider_earth_2017, schneider_harnessing_2023, Eyring2024, lam2022graphcast, bi2023accurate, kochkov2023neural}. Such methods seek to inherit the desirable properties of both numerical and ML models, while attenuating their drawbacks. A group of methods in this category focuses on correcting the dynamics \textit{on-the-fly} by \textit{intrusively} modifying classical numerical models \cite{arcomano_hybrid_2022, clark_correcting_2022, sanderse2024scientific, Boral_NiLES:2023}. The underlying dynamical model provides a strong inductive bias, which reduces the training data requirements compared to purely-data driven models, and helps capture many of the dynamical properties of the system \cite{kochkov_machine_2021,dresdner2022learning}.
In the context of climate modeling, recent approaches seek to learn state-dependent closure terms for the effect of the unresolved sub-grid-scale processes on the resolved scales. Such approaches have been shown to be effective in both reducing overall bias \cite{watt-meyer_correcting_2021, Guillaumin2021} as well as capturing unresolved processes \cite{arcomano_hybrid_2023}. Furthermore, they have been demonstrated on a range of systems ranging from idealized aquaplanet configurations \cite{yuval_stable_2020,rasp_deep_2018,brenowitz_spatially_2019,yuval_use_2021, iglesias-suarez2024} to more realistic global climate models \cite{bora_learning_2023,bretherton_correcting_2022}.

However, these intrusive ML corrections have several drawbacks. Their implementation requires integration into the original dynamical model, which can be a complex process \cite{mcgibbon_fv3gfs-wrapper_2021}. Furthermore, these closure terms are typically learned offline, without interaction with the dynamical system, since very few dynamical models meet the fast differentiation requirements for integrated online learning \cite{kochkov2023neural}. In addition, although advances have been made in stabilizing such hybrid models, long-term instability can still be an issue \cite{zhang_error_2021,wikner_stabilizing_2022,yuval_use_2021}. Gradient-free ensemble Kalman methods have recently been proposed that enable online learning in hybrid systems \cite{lopezgomez2022, Christopoulos_2024}. These methods can learn from long-term statistics to guarantee stability, but their application is limited to relatively sparse ML corrections.

Another group of hybrid methods focuses on machine learning \textit{non-intrusive} corrections, meant to be applied as a post-processing step. Since there is no interaction with the numerical solver, these methods are long-term stable by design. Post-processing methods apply a machine learned map to biased trajectories of the dynamical system such that the statistics of the output match those of the training data. The need to train on statistics rather than trajectories is necessitated by the chaotic nature of the underlying system and the absence of paired (or aligned) data available for training. Such techniques have been applied to coarse resolution weather and climate simulations in the context of statistical debiasing \cite{blanchard_multi-scale_2022,mcgibbon_global_2023, Li2024} and downscaling \cite{Vandal_2017:DeepDS,wan_debias_2023,Wilby_1998:downscaling}. In the context of debiasing, multiple methods have been explored in the literature including generative models based on optimal transport theory \cite{arbabi_generative_2022}, temporal-convolutional-network (TCN) and LSTM networks \cite{blanchard_multi-scale_2022}, generative adversarial networks (GAN) \cite{mcgibbon_global_2023}, unsupervised image-to-image networks (UNIT) \cite{fulton_bias_2023}, and diffusion models \cite{Li2024}. However, the requirement to reproduce the statistics of the training data greatly limits the potential of such methods to generalize to longer trajectories than those observed in training. 

To tackle such limitation, we propose post-processing methodology to debias coarse-resolution climate simulations that is able to correct statistics of rare extreme events even when these have return periods far longer than the period spanned by the training dataset. Our proposed methodology seeks to extend the application of trajectory-based post-processing methods to long-time simulations through the use of probabilistic neural network models trained on specific paired sets of training data. Specifically, our framework leverages a recently developed methodology to generate paired climate trajectories \cite{barthel_sorensen_non-intrusive_2024,zhang_machine_2024} that avoids common pitfalls of training ML algorithms for chaotic systems. These paired trajectories are then used to learn a probabilistic post-processing operator using variational inference methods.

Variational inference methods seek to approximate a distribution using its samples by solving an optimization problem where the distribution itself is parameterized by a neural network. In this case, we leverage Variational Auto-Encoders (VAEs) \cite{kingma_auto-encoding_2022} coupled with Long-Short-Term-Memory based recurrent neural network (RNN) architectures \cite{hochreiter_long_1997} and ensemble learning \cite{opitz1999popular}. VAEs compress the system's state into a probabilistic latent representation whose distribution is learned variationally. RNNs map one trajectory to another by processing snapshots sequentially using a latent representation of the current and previous states. By replacing the latent representation in RNNs by a probabilistic one learned variationally one obtains a map from a trajectory to a distributions of \textit{plausible} trajectories. \textcolor{black}{Furthermore, we train a small ensemble of such networks using the same data and different random seeds. Thus, the final algorithm defines a map from a trajectory to a \textit{composite distribution of trajectories}, which captures the uncertainty of the system more accurately, in contrast to deterministic models that tend to learn the expectation.}

\textcolor{black}{Our variational extension greatly increases the generalization and extrapolation capabilities of deterministic models used in previous work \cite{barthel_sorensen_non-intrusive_2024}. This allows us to accurately predict the probability of tail-risk events with longer return periods than the training period, and which are therefore likely to be missing entirely from the training data. Furthermore, we illustrate the advantages of our framework on a range of metrics not previously considered, including two-point correlations, regional variation, and extreme event statistics. We also conduct a systematic comparison of several variational architectures to serve as a guide to researchers looking to implement our framework.} In summary, our approach bypasses the three main difficulties encountered by many ML-based surrogates for chaotic systems, namely: long time inference stability, generalizability, and training stability. Our approach is stable for indefinitely long time horizons by construction, sample efficient, easy to implement, and empirically able to extrapolate statistically relevant properties. \textcolor{black}{In this work we apply our methodology on an anisotropic 2D quasi-geostrophic flow, which, albeit simple, captures many of the core difficulties of models with more complex physics.} Crucially, it can be simulated over very long time horizons at reasonable computational cost. This last property allows us to study the behavior of very long trajectories, which is infeasible with the time-horizon of current climate datasets.

The remainder of this article is organized as follows. In \S\ref{sec:Math} we outline the mathematical formulation of problem under investigation and in \S\ref{sec:QG} we introduce the specific prototypical climate model to be analyzed. \S\ref{sec:ML} summarizes the specific machine learning architectures we employ, and our results are presented in \S\ref{sec:Results}. We conclude with a discussion of the implications of our results in \S\ref{sec:Disc}.

\section{Mathematical Framework}\label{sec:Math}
We consider a discretized representation of an ergodic chaotic dynamical system
\begin{equation}\label{eq:fine_model}
 \partial_t \mathbf{q}=F(\mathbf{q}), \ \  \mathbf{q}\in\mathbb{R}^N,\ 
\end{equation}
with initial conditions $\mathbf{q_0}$ following a pre-defined distribution $\mu_0$, which in turn induces a distribution of trajectories. Here we loosely define a chaotic system as one whose trajectories are highly \textit{sensitive to perturbations of initial conditions.} Specifically, chaotic systems are characterized by having a positive Lyapunov exponent: small discrepancies in the initial conditions are exaggerated exponentially over time \cite{strogatz2018nonlinear}.
In defining the system \eqref{eq:fine_model} we assume $N$ is large enough that the statistics of the solution $\mathbf{q}$ do not change with increasing $N$ -- we refer to such a system as being \textit{``fully-resolved''}. 
Correspondingly, we also consider an \textit{``under-resolved''} discretization of the same dynamical system, described by 
\begin{equation}\label{eq:coarse_model}
\partial_t \mathbf{v}= f(\mathbf{v}), \ \  \mathbf{v}\in\mathbb{R}^n,
\end{equation}
where $n<N$, and, crucially, the statistics of $\mathbf{v}$ depend on $n$. Finally, we define the projection of the fully-resolved solution onto the coarse grid via the projection operator $\mathbf{P}$
\begin{equation} \label{eq:RD}
	\mathbf{u}=\mathbf{P}\mathbf{q},\ \   \mathbf{u} \in \mathbb R^n.
\end{equation}
Moving forward, $\mathbf{u}$ will be referred to as the reference data (RD) and $\mathbf{v}$ will be referred to as the coarse data (CR).
We also consider the discretization in time of the solutions of \eqref{eq:fine_model} and \eqref{eq:coarse_model} to snapshots sampled equi-spaced in time, resulting in the sequences $\{\mathbf{v}_j\}_{j=1}^T$ and $\{\mathbf{u}_j\}_{j=1}^T$, where $\mathbf{u}_j = \mathbf{P}\mathbf{q}_j$. 

The objective of this work is to learn a \textit{parametric correction operator}
\begin{equation}\label{eq:G_operator}
    \mathcal{G}_{\theta} : \mathbb{R}^{n \times T } \rightarrow  \mathcal{P}( \mathbb{R}^{n \times T }),
\end{equation}
where $T$ is the length of the trajectories, $\mathcal{P}( \mathbb{R}^{n \times T})$ is the push-forward map by $\mathbf{P}$ of a distribution of trajectories of system \eqref{eq:fine_model}, and $\theta$ are the parameters of the map. Thus, $\mathcal{G}_{\theta}$ maps trajectories from the distribution of the under-resolved (coarse) system \eqref{eq:coarse_model} to distributions of trajectories of the projected fully-resolved (reference) system \eqref{eq:fine_model}. We are focused on the statistical evaluation of long term climate risks, and thus the aim of \eqref{eq:G_operator} is not to approximate any specific reference trajectory on a snapshot-by-snapshot basis, but rather to generate plausible trajectories which reflect the statistics of the reference data.

We highlight that the operator $\mathcal{G}_{\theta}$ maps trajectories from $n$-dimensional state space to $n$-dimensional state space, and is not intended to recover the fine scales unresolved by the coarse model. Therefore, all results presented in this work should be understood as being defined on the coarse grid.

\subsection{Training on Nudged Simulations}
The primary obstacle to learning a map $\mathcal{G}_{\theta}$ is that the systems associated to $\mathbf{v}$ and $\mathbf{u}$ are chaotic, and therefore there is no natural pairing between trajectories \cite{wan2023debias}. One could learn a map between any arbitrary pair of trajectories, but such map will be highly specific to that particular ordering, and in general will not generalize to unseen data. In addition, for the sake of generalization the mapping must directly encode the spatiotemporal dynamics of the system \eqref{eq:fine_model}, not just the statistics of the specific trajectories used in training. 
This additional constraint stems from the downstream application: practical long-term (multi-century) climate forecasting will require training correction operators on the few decades of available high quality data whose statistics are not converged -- especially for rare events whose characteristic return period is on the order of centuries. If $\mathcal{G}_{\theta}$ is trained to simply generate trajectories drawn from the distribution defined by the training data such extrapolation is often impossible without additional strong inductive biases, which themselves are usually not well defined.

To overcome these challenges, we employ the framework introduced by \citeA{barthel_sorensen_non-intrusive_2024} in which the correction operator is trained on trajectory pairs consisting of a fully-resolved reference trajectory and an under-resolved trajectory \textit{nudged} towards that reference trajectory. We briefly summarize the mathematical rationale of the approach below, and refer the interested reader to \citeA{barthel_sorensen_non-intrusive_2024} for a more detailed presentation. 

Consider the deviation between the under- and fully- resolved representations of the dynamical system
\begin{equation}\label{eq:q_def}
\boldsymbol{\delta} \equiv \mathbf{v}-\mathbf{u}, \ \ \ \boldsymbol{\delta}\in\mathbb{R}^n,
\end{equation} 
which is governed by the system
\begin{equation} \label{eq:q_model}
    \partial_t \boldsymbol{\delta} = f(\boldsymbol{\delta}+\mathbf{ P} \mathbf{q})-\mathbf{P}F(\mathbf{q}).
\end{equation}
Due to the chaotic nature of the system, $\boldsymbol{\delta}$ will grow exponentially. This is known as chaotic divergence and makes a map between any two arbitrary realizations of $\mathbf{v}$ and $\mathbf{u}$ meaningless. This divergence can be constrained through the introduction of a small damping term on the right hand side of \eqref{eq:q_def} resulting in
\begin{equation}
    \partial_t \boldsymbol{\delta}_{\tau} = f(\boldsymbol{\delta}_{\tau}+\mathbf{P} \mathbf{q})-\mathbf{P}F(\mathbf{q})-\frac{1}{\tau}\boldsymbol{\delta}_{\tau},
\end{equation}
which when expressed in terms of the original variables takes the form
\begin{equation}\label{eq:nudged_model}
    \partial_t \mathbf{v}_{\tau} =f(\mathbf{v}_\tau)-\frac{1}{\tau}(\mathbf{v}_{\tau}-\mathbf{u}), \ \  \mathbf{v}_{\tau}\in\mathbb{R}^n.
\end{equation}
If the reference solution $\mathbf{u}$ is known, the system \eqref{eq:nudged_model} is said to be \textit{nudged} towards $\mathbf{u}$ -- an approach which originates in the field of data assimilation, where it has been used to improve the predictive capabilities of weather models \cite{huang_development_2021,miguez-macho_regional_2005,storch_spectral_2000,sun_impact_2019} . The forcing term on the right hand side of \eqref{eq:nudged_model} is known as the nudging tendency, and the user-defined constant $\tau$ represents a time scale over which this forcing acts. The nudging tendency will have a negligible effect when $(\mathbf{v}_{\tau}-\mathbf{u})$ is small and an $O(1)$ effect on the dynamics only when the deviation $(\mathbf{v}_{\tau}-\mathbf{u})$ grows to be $O(\tau)$. Through a multiscale analysis, \citeA{barthel_sorensen_non-intrusive_2024} showed that nudging is equivalent to forcing the dynamics evolving on time scales slower than $\tau$ to follow the slow dynamics of the reference trajectory $\mathbf{u}$, while the faster dynamics are free to evolve according to the unforced coarse dynamics \eqref{eq:coarse_model}.

\begin{figure}
    \centering
    \includegraphics[trim = 0 0 0 0, width=0.95\textwidth]{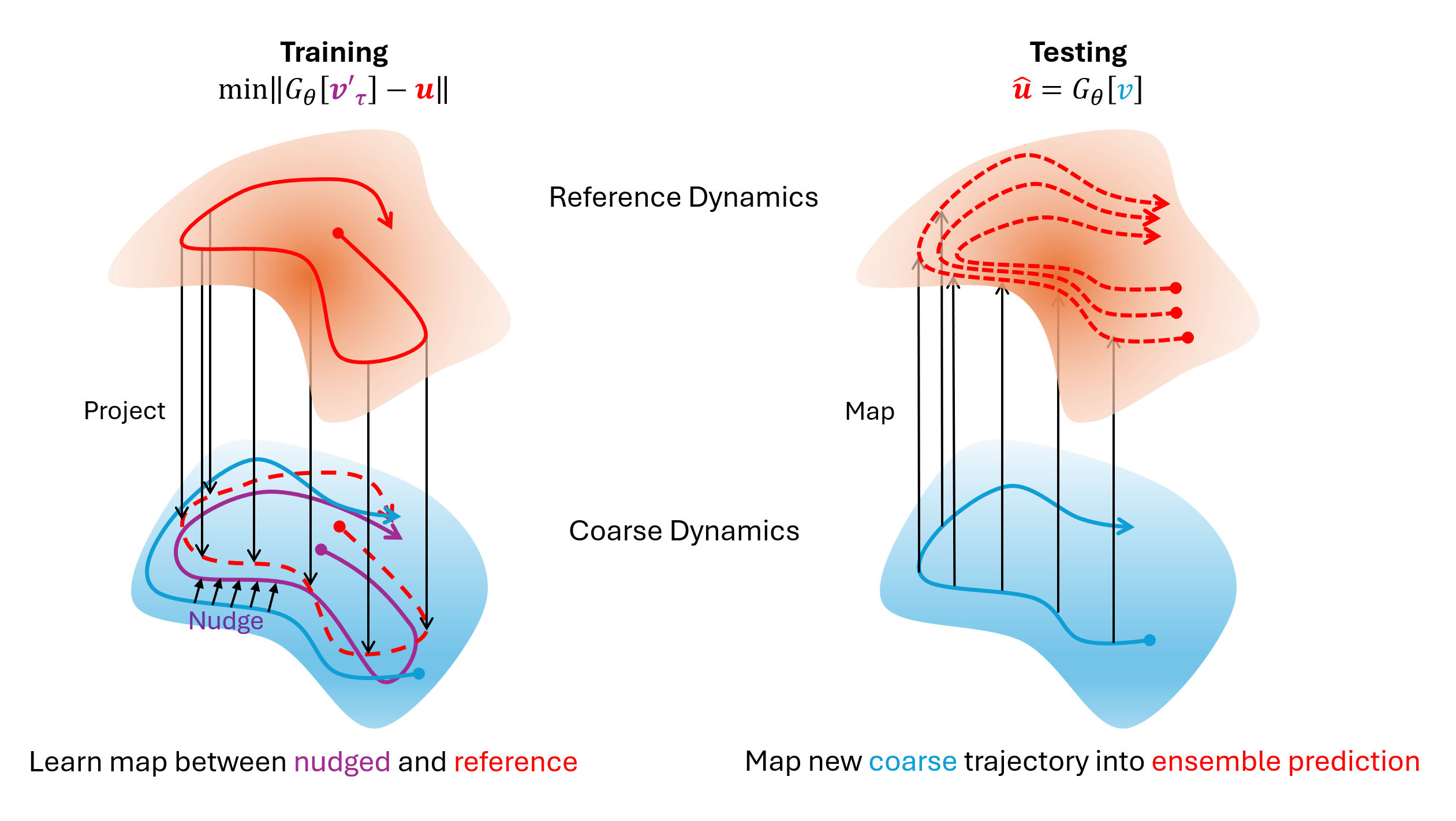}
    \caption{Diagram of the nudging-based machine learning framework.}
    \label{fig:diagram}
\end{figure}

Training on the pair of trajectories $\mathbf{v}_{\tau}$ and $\mathbf{u}$ allows the correction operator $\mathcal{G}_{\theta}$ to learn the fast dynamics of the fully-resolved system which are most affected by the lack of resolution, while being minimally corrupted by the chaotic divergence of the large-scale slow dynamics. The aim therein is to learn a map which reliably maps trajectories in the distribution induced by the coarse dynamics \eqref{eq:coarse_model} to the distribution induced by the reference (fully-resolved) dynamics \eqref{eq:fine_model}. However, the inclusion of the nudging tendency in \eqref{eq:nudged_model} introduces artificial dissipation, which causes the spectrum of the nudged solution $\mathbf{v}_{\tau}$ to differ from that of the free running solution $\mathbf{v}$. To address this, we define the spectrally corrected nudged solution
\begin{equation}\label{eq:spectral_correction}
    \mathbf{v}_{\tau}'=\mathcal{F} ^{-1}[a_k \hat{\mathbf{v}}_{\tau,k}],
\end{equation}
where $\hat{\mathbf{v}} \equiv \mathcal{F}[\mathbf{v}]$ is the spatial Fourier transform and $a_{k}$ is the spectral ratio defined as
\begin{equation}
    a_{k} \equiv \sqrt{\int_0^T|\hat{\mathbf{v}}_{k}|^2dt \left(\int_0^T|\hat{\mathbf{v}}_{\tau, k}|^2dt \right)^{-1}} .
\end{equation}
We note that several other strategies to address such spectral inconsistencies have been proposed such as 4DVar \cite{dimet_variational_1986,mons_reconstruction_2016,wang_discrete_2019} or ensemble variational methods \cite{liu_ensemble-based_2008,mons_reconstruction_2016, buchta_observation-infused_2021}. We utilize the simple spectral correction due to its ease of implementation and the fact that it does not require iterative simulation of the governing equations as some of these other methods. In practice the training data consists of 3 trajectories, the reference data $\mathbf{u}$, the spectrally-corrected nudged coarse data $\mathbf{v}'_{\tau}$, and a free running coarse trajectory $\mathbf{v}$ used for the spectral correction \eqref{eq:spectral_correction}. We then formulate the general supervised learning problem
\begin{equation}\label{eq:learning_prob}
     \min_{\theta} \int_0^T \left\|\mathcal{G}_{\theta}[\mathbf{v}_\tau']-\mathbf{u}\right\|^2dt,
\end{equation}
where $\mathbf{v}_\tau'$ and $\mathbf{u}$ are understood to be discrete trajectories. By formulating the learning in terms of trajectories -- and not just statistics -- the learned map directly encodes the temporal dynamics of the system. This allows for the possibility of the learned map to extrapolate to trajectories which are much longer than the training data which would be impossible if $\mathcal{G}_{\theta}$ was trained only to reproduce the statistics of the data seen in training \cite{blanchard_multi-scale_2022}. A diagram of the general learning framework is shown in Figure \ref{fig:diagram}.

\section{Quasi-Geostrophic Model}\label{sec:QG}
Similarly to \cite{barthel_sorensen_non-intrusive_2024}, we consider a two-layer quasi geostrophic model as prototypical climate model. The model is defined on 2D doubly periodic Cartesian grid, $(x,y) \in [0,2\pi]^2$, and takes the form
\begin{equation}\label{eq:QG_model}
\frac{\partial q_j}{\partial t} +\left(U_j +  \hat{\mathbf{k}} \times \nabla \psi_j  \right) \cdot\nabla q_j + \left(\beta +k_d^2U_j\right)\frac{\partial \psi_j}{\partial x} = -\delta_{2,j}r\nabla^2\psi_j - \nu \nabla^8q_j,
\end{equation}
where $j = 1, 2$ corresponds to the upper and lower layers. The dependent variable appears in two forms: $q_j(x,y,t)$ and $\psi_j(x,y,t)$, which are the potential vorticity and stream function respectively. Without loss of generality, all results in this work will be presented in the form of the stream functions $\psi_j$. 

The system is parameterized by the bottom-drag coefficient $r$, the beta-plane approximation parameter $\beta$, and the deformation frequency $k_d^2$. In this work we fix $[r,\beta,k_d^2] = [0.1,2.0,4.0]$ -- values consistent with mid-latitude atmospheric flow. The imposed zonal mean flow is given by $U_j = -1^{(j+1)}U$, with $U=0.2$. 

\begin{figure}
        \centering
        \begin{tabular}{ll}
       ~\  ~\  ~\ ~\ ~\ 
        \includegraphics[trim =  0 0 0 0, width=0.85\textwidth]{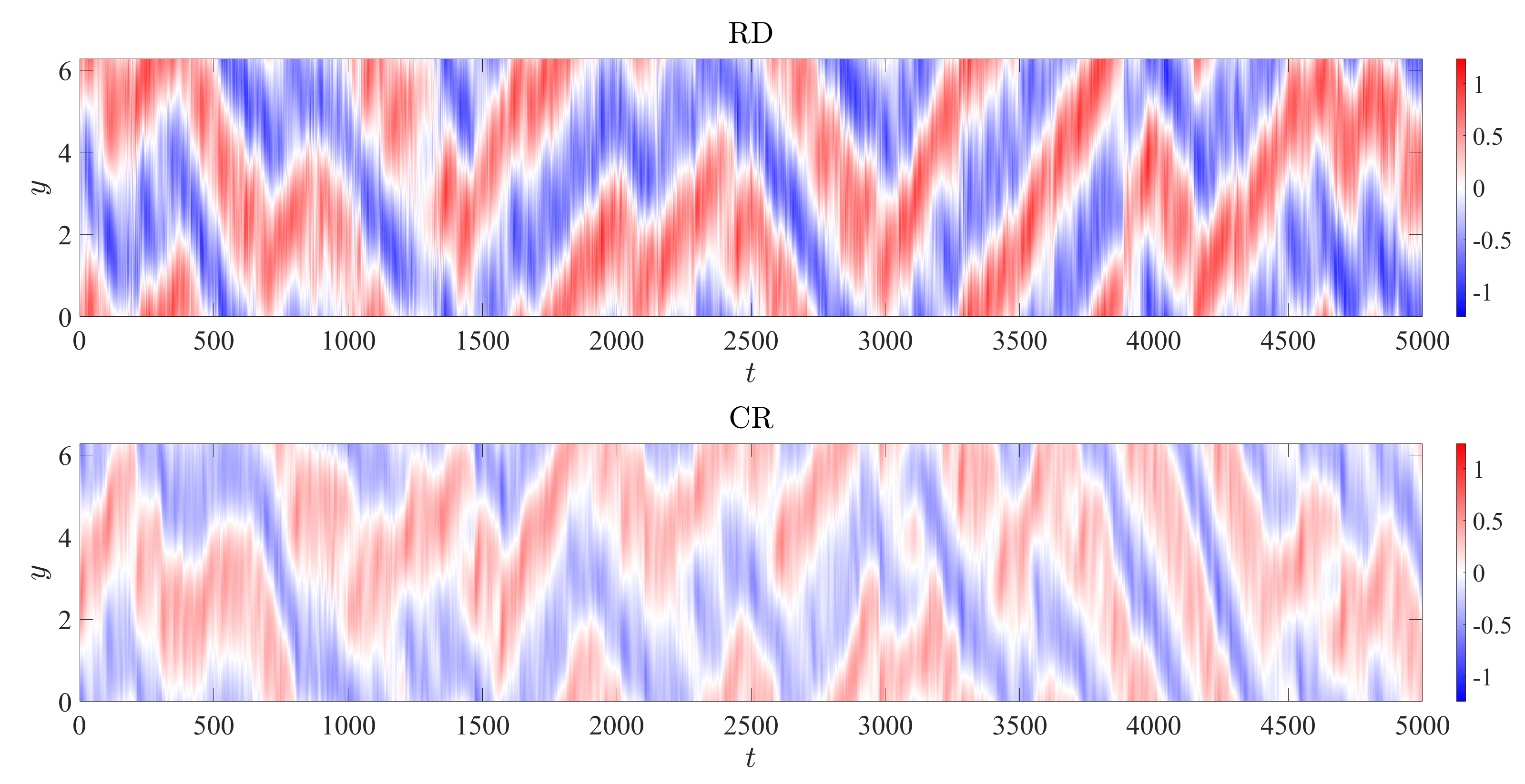} \\
          ~\  ~\   ~\    ~\  ~\  ~\  ~\  ~\  ~\    ~\   ~\  ~\   ~\   ~\  ~\  ~\  ~\   ~\   ~\   ~\  ~\  ~\  ~\  ~\  ~\  ~\  ~\  ~\  ~\  ~\  (a)  \\
        \includegraphics[trim =  0 0 0 0, width=0.5\textwidth]{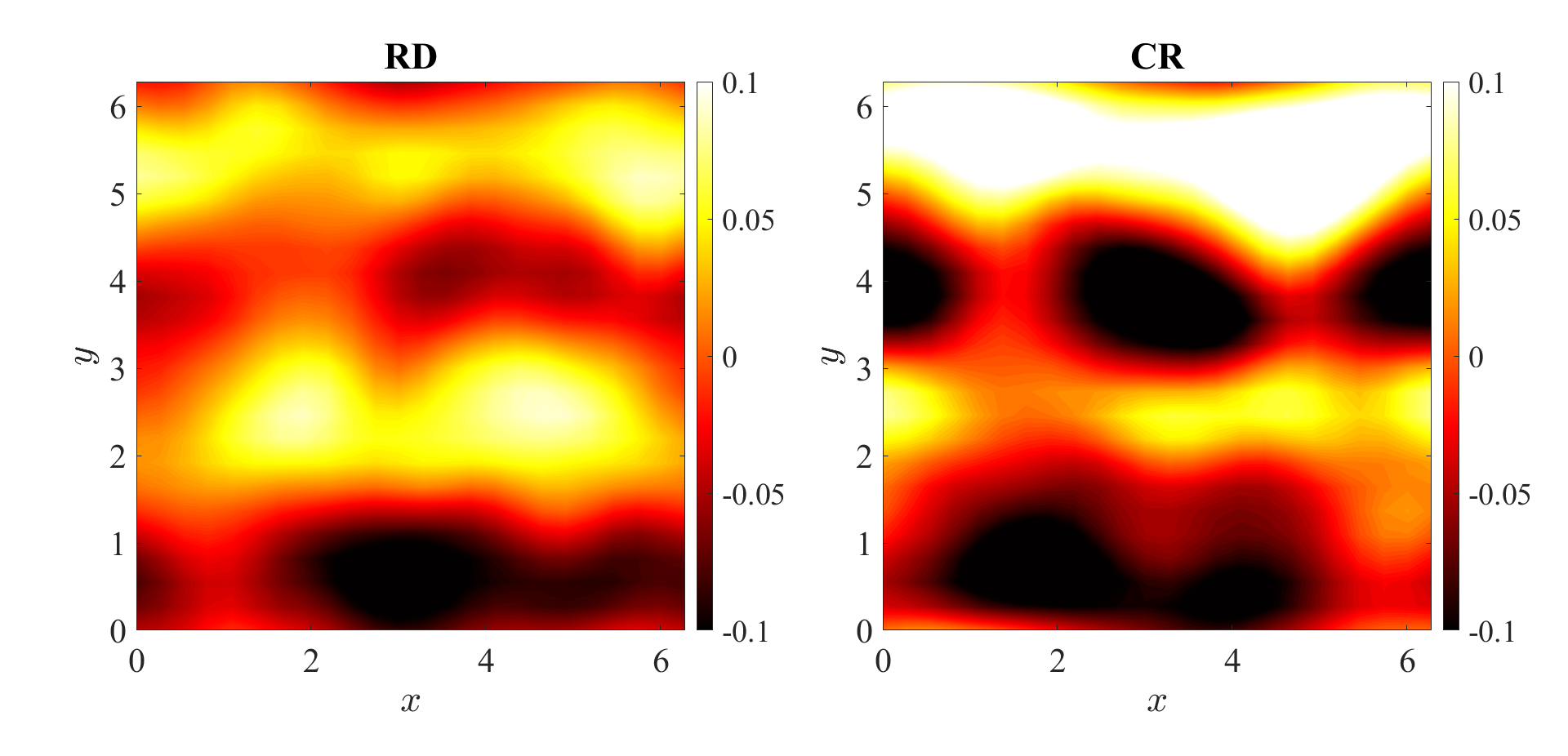} 
        \includegraphics[width=0.5\textwidth]{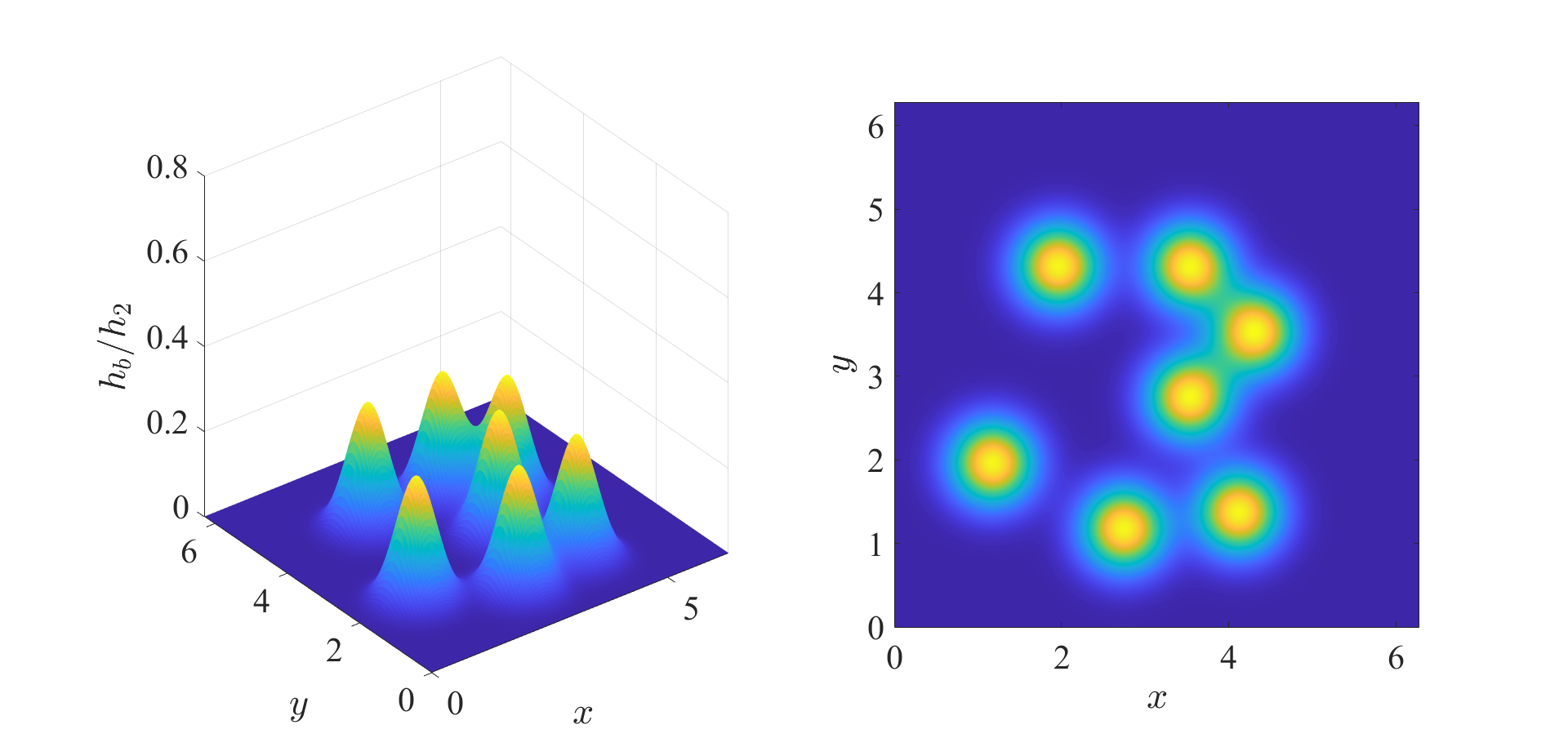}
        \\
          ~\  ~\    ~\   ~\   ~\   ~\  ~\  ~\  ~\   ~\   ~\   ~\  ~\  ~\  ~\   ~\   ~\  (b)  ~\  ~\  ~\  ~\  ~\  ~\ ~\  ~\  ~\  ~\  ~\  ~\  ~\  ~\  ~\  ~\  ~\  ~\  ~\  ~\  ~\  ~\ ~\  ~\  ~\  ~\  ~\  ~\  ~\  ~\  ~\  ~\  ~\  (c)   \\
        
        \end{tabular}
        \caption{Zonal average (a) and normalized covariance (b) of the lower layer stream function $\psi_2$ of the RD and CR data sets. Illustration of the bottom topography profile (c).}
        \label{fig:QG_example}
\end{figure}

To quantify the effectiveness of our methodology to anisotropic problems we introduce topography on the bottom surface. The topography profile $h_b(x,y)$ is introduced through the definition of the potential vorticity
\begin{equation}\label{eq:vort_def}
q_j = \nabla^2 \psi_j + \frac{k^2_d}{2}\left(\psi_{3-j} - \psi_j\right) + \frac{f_0}{h_2} h_b(x,y) \delta_{j,2}.  
\end{equation}
Here $f_0$ is the inertial frequency which we set to 1, $h_2$ is the thickness of the lower layer, and $\delta_{j,2}$, indicates that the topography term is only included in the definition of the lower layer potential vorticity $q_2$. We consider a topography profile consisting of seven randomly spaced Gaussians with equal variance
\begin{equation}\label{eq:topo}
    h_b(x,y) = A \sum_{j=1}^7 e^{-\frac{(x-a_j)^2 + (y-b_j)^2}{\sigma^2}},
\end{equation}
where the coordinates $[a_j,b_j]$ and variance $\sigma^2$ represent the centers and width of the Gaussian ``mountains''. The specific values 

were chosen to ensure that the profile would not violate the periodic boundary conditions. 
An illustration of the topography profile is shown in Figure 
\ref{fig:QG_example}c.

Equations \eqref{eq:QG_model} and \eqref{eq:vort_def} are solved using a spectral method in space
and then integrated using a $4^{th}$ order Runge-Kutta scheme in time. We consider $128\times128$ and $24\times24$ grid to represent the specific fully- and under- resolved  systems \eqref{eq:fine_model} and \eqref{eq:coarse_model}, respectively. For each case, we run a \textit{single} simulation for 35,000 time units, the first 1,000 time units are used for training, and the remaining 34,000 are used for testing.
One additional nudged simulation over 1000 time units is performed to generate the training data \eqref{eq:spectral_correction} needed to construct the supervised learning problem \eqref{eq:learning_prob}.

Figure \ref{fig:QG_example}a shows the zonally averaged flow field as an illustrative example. Note the difference in amplitude between the RD and CR solutions. Figure \ref{fig:QG_example}b shows the spatial variation of the normalized variance of the stream function data defined as
\begin{equation}
    \tilde{\sigma}(x,y) = \frac{\sigma(x,y) - \overline{\sigma}}{\overline{\sigma}},
\end{equation}
where the variance is computed over the temporal dimension (34,000 time units) and $\overline{\sigma}$ denotes a spatial average. This highlights both the anisotropy present in the flow as well as the non-trivial differences in the spatiotemporal features of the RD and CR data sets. Finally, we reemphasize that the RD dataset represents the high resolution solution projected onto the coarse grid, and thus all data and results shown in this work are defined on the coarse $24\times 24$ grid.

\section{Machine Learning Architecture}\label{sec:ML}
Here we provide a brief description of the neural network architectures and uncertainty quantification strategies investigated in this work. 
We reemphasise that the aim of the current approach is to train correction operators which are effective when applied to unseen trajectories which are significantly (perhaps orders of magnitude) longer than the training trajectories.
To this end, we investigate three probabilistic extensions of the previously validated Long Short Term Memory (LSTM)-based network \cite{barthel_sorensen_non-intrusive_2024}, all based on the principle of VAEs \cite{kingma_auto-encoding_2022}. To illustrate the number of possible, and often subtle, interactions between the VAE and LSTM we begin with a brief outline of a basic RNN and then explain how each of the three architectures under investigation builds upon this baseline.
At a high level, the VAE introduces a probabilistic latent space which in theory allows the network to learn embeddings of the limited training data in a manner which is cognizant of and robust to the limitations of that data. The primary variation we investigate here is whether this latent space is implemented \textit{``upstream''} or \textit{``downstream''} of the LSTM unit in the computational graph of the network as a whole. Much of the discussion in \S\ref{sec:Results} and \S\ref{sec:Disc} focuses on the advantages and disadvantages of each and how these may be exploited or mitigated respectively.
\subsection{Recurrent Neural Networks}
One of the most widely used class of ML architectures for modeling temporal sequences such as the climate systems which motivate our research is the RNN \cite{graves_unconstrained_2007,sutskever_recurrent_2008,graves_speech_2013,graves_generating_2014,sutskever_sequence_2014,cho_learning_2014}. An RNN layer transforms the input sequence $\mathbf{x}=[\mathbf{x}_1,...,\mathbf{x}_T], \mathbf{x}_t \in \mathbb{R}^n$ into an output $\mathbf{y}=[\mathbf{y}_1,...,\mathbf{y}_T], \mathbf{y}_t \in \mathbb{R}^m$, via a hidden state $\mathbf{h}=[\mathbf{h}_1,...,\mathbf{h}_T], \mathbf{h}_t \in \mathbb{R}^d$ according to the following recursive push-forward equations
\begin{align}\label{eq:rnn_basic}
     &\mathbf{h}_t = f_h \left(\mathbf{U}\mathbf{x}_t + \mathbf{W}\mathbf{h}_{t-1} + \mathbf{b} \right),\\
    &\mathbf{y}_t = f_o \left(\mathbf{V}\mathbf{h}_t + \mathbf{c}  \right), 
\end{align}
where $\mathbf{U}, \mathbf{W}, \mathbf{V}, \mathbf{b}, \mathbf{c}$ represent the trainable parameters, and both $f_h$ and $f_o$ are the generally nonlinear activation functions. A graphical representation of the basic RNN unit is given in Figure \ref{fig:dssm}a. This basic formulation is generally augmented using gating mechanisms which alleviate the problem of vanishing gradients \cite{pascanu2013difficulty} during training which arise due to exponentially small weights assigned to long term dependencies. Specifically, all of the network architectures explored in this work are built on LSTM unit \cite{hochreiter_long_1997}, as LSTM based architectures have generally demonstrated superior ability to capture long time dependencies as compared to other designs such as the Gated Recurrent Unit (GRU) \cite{cho_properties_2014}.


\subsection{Variational Auto Encoder}
The ML correction operator will generally encounter many events which were rarely or not at all seen in training. One architecture that has been proposed to enable such generalization (for non-sequential data) is the Variational Auto-Encoder (VAE) \cite{kingma_auto-encoding_2022}.  A standard Auto-Encoder (AE) is a type of data compression architecture which projects the input data $\mathbf{x}$ onto a reduced order latent space $\mathbf{z}$ and then expands it back to an approximation of the original input data $\tilde{\mathbf{x}}$. The AE is then generally trained to minimize the reconstruction error: $\|\mathbf{x}-\tilde{\mathbf{x}}\|$. The VAE replaces the deterministic latent space in the standard AE with a probabilistic latent space, where for each forward pass the latent space representation is sampled from a distribution, which for ease of parameterization, is generally assumed to be Gaussian $\mathbf{z} \sim \mathcal{N}\left(\boldsymbol{\mu}_z,\boldsymbol{\sigma}_z \right)$. From an implementation point of view, this implies that each embedding is now not just a single number but a mean and a variance. This extension to a latent space of distributions regularizes or smooths out the latent space ensuring that that structures which are similar in physical space will have similar embeddings -- a property which is not guaranteed in a deterministic encoder-decoder network. This built in uncertainty improves the extrapolation capabilities of the network by increasing the likelihood that \textit{never-before-seen} structures will be encoded into latent space representations which are similar to the embedding of similar structures which \textit{were seen} in training, and thereby increasing the likelihood of an accurate decoding.

However, for this framework to be useful some regularization constraints are required on the latent space distribution. For example, without constraints, the network is liable to over-fit to the training data and converge to a latent space whose mean values are distant from one another and whose covariances vanish thereby negating the benefit of the probabilistic framework entirely. This regularization is achieved through an addition to the loss function which penalizes deviations of the distribution $p(z) \sim \mathcal{N}\left(\boldsymbol{\mu}_z,\boldsymbol{\sigma}_z \right)$ from a standard Normal distribution: $\mathcal{N}\left(\mathbf{0}, \mathbf{I}\right)$. We note that while other priors are possible, these were not pursued in this work.

\subsection{Probabilistic Recurrent Neural Networks}
The probabilistic treatment of sequential temporal data requires the combination of the RNN and VAE frameworks. Such hybrid architectures are also known as Deep State Space Models (DSSMs) \cite{gedon_deep_2021}, however to minimize unnecessary jargon we will refer to such models simply as probabilistic (as opposed to deterministic) RNNs. 

Here we investigate three recently proposed probabilistic RNN architectures: the VAE-RNN \cite{fraccaro_sequential_2016,fraccaro_deep_2018}, the stochastic RNN (STORN) \cite{bayer_learning_2015}, and the variational RNN (VRNN) \cite{chung_recurrent_2016}. 

\subsubsection{VAE-RNN}
The VAE-RNN is the simplest form of probabilistic RNN. In this case a VAE is simply appended to the output of the RNN at each time step independently -- this is illustrated graphically in figure \ref{fig:dssm}b. The recursive push-forward equations for the VAE-RNN are
\begin{align}\label{eq:rnn_vae_rnn}
     &\mathbf{h}_t = f_h \left(\mathbf{U}\mathbf{x}_t + \mathbf{W}\mathbf{h}_{t-1} + \mathbf{b} \right),\\
     &\mathbf{z}_t \sim  \mathcal{N}\left(\boldsymbol{\mu}_z(\mathbf{h}_t),\boldsymbol{\sigma}_z(\mathbf{h}_t) \right),\\ 
    &\mathbf{y}_t = f_o \left(\mathbf{V}\mathbf{z}_t + \mathbf{c}  \right), 
\end{align}
where $\boldsymbol{\mu}^2_z(\mathbf{h}_t) = \mathbf{B}\mathbf{h}_t$ and $\boldsymbol{\sigma}_z(\mathbf{h}_t) = \textrm{softplus}\left(\mathbf{C}\mathbf{h}_t\right)$ are both themselves parameterized through the trainable weight matrices $\mathbf{A}$ and $\mathbf{B}$ and the use of the $\textrm{softplus}$ activation function ensures a positive variance. A critical (and limiting) feature of the VAE-RNN architecture is that the latent space dependency is \textit{downstream} of the recurrence relationship and thus there is no communication between time steps $\mathbf{z}_t$. The following two architectures remedy this limitation. 

\subsubsection{STORN}
\begin{figure}
    \centering
    \includegraphics[width=0.95\textwidth]{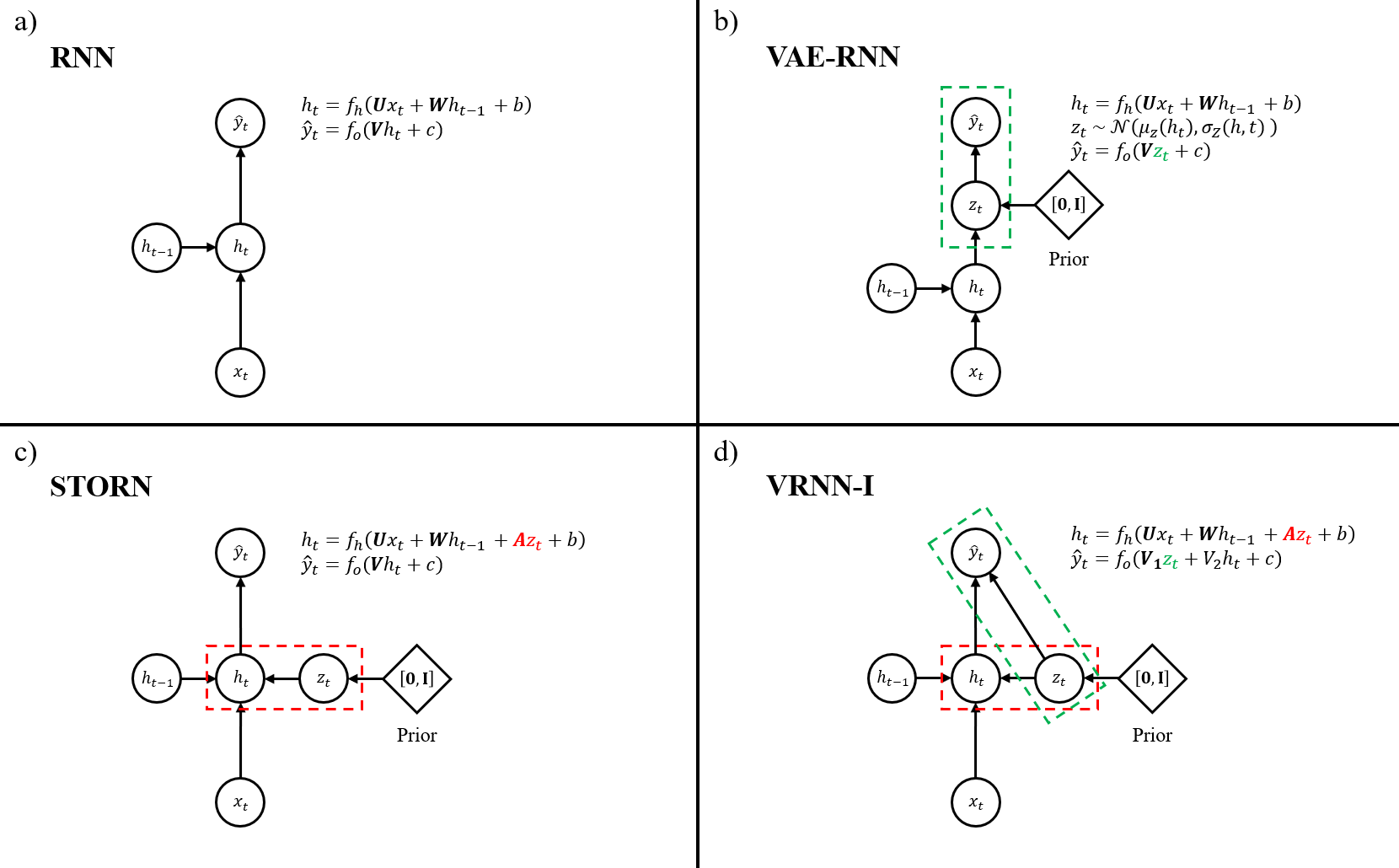}
    \caption{Graphical model and recursive evaluation equations for the four network architectures considered in this work: basic RNN (a), VAE-RNN (b), STORN (c), VRNN (d). Latent space dependencies upstream and downstream of the recurrent layer are marked red and green respectively.}
    \label{fig:dssm}
\end{figure}

The STORN architecture does not append a VAE to the output of the RNN but instead introduces the latent space upstream of the recurrence relationship, namely as an additional input to the RNN. Specifically, it consists of the following push forward equations
\begin{align}\label{eq:rnn_storn}
 &\mathbf{z}_t \sim  \mathcal{N}\left(\boldsymbol{\mu}_z(\mathbf{x}_t),\boldsymbol{\sigma}_z(\mathbf{x}_t) \right),\\ 
     &\mathbf{h}_t = f_h \left(\mathbf{U}\mathbf{x}_t + \mathbf{W}\mathbf{h}_{t-1} + \mathbf{A}\mathbf{z}_t + \mathbf{b} \right),\\
    &\mathbf{y}_t = f_o \left(\mathbf{V}\mathbf{h}_t + \mathbf{c}  \right),
\end{align}
where the latent space is parameterized in terms of the input variable $\boldsymbol{\mu}^2_z(\mathbf{x}_t) = \mathbf{B}\mathbf{x}_t$ and $\boldsymbol{\sigma}_z(\mathbf{x}_t) = \textrm{softplus}\left(\mathbf{C}\mathbf{x}_t\right)$. A graphical illustration of the basic STORN architecture is given in figure \ref{fig:dssm}c. 
\subsubsection{VRNN}
The VRNN architecture includes both an upstream and downstream latent space dependency, and can
be interpreted as a combination of the VAE-RNN and STORN architectures. The latent space is introduced as an input to the RNN but is also appended to its output. The generative equations are
\begin{align}\label{eq:rnn_vrnn}
 &\mathbf{z}_t \sim  \mathcal{N}\left(\boldsymbol{\mu}_z(\mathbf{x}_t),\boldsymbol{\sigma}_z(\mathbf{x}_t) \right)\\ 
     &\mathbf{h}_t = f_h \left(\mathbf{U}\mathbf{x}_t + \mathbf{W}\mathbf{h}_{t-1} + \mathbf{A}\mathbf{z}_t + \mathbf{b} \right)\\
    &\mathbf{y}_t = f_o \left(\mathbf{V}_1\mathbf{h}_t + \mathbf{V}_2 \mathbf{z}_t +  \mathbf{c}  \right) \label{eq:vrnn_decoder}
\end{align}
where the latent space is parameterized as in the STORN model. \citeA{chung_recurrent_2016} investigate both a standard Gaussian prior (VRNN-I) as well as a generally time dependent prior which is learned during the training phase. Here we consider only the VRNN-I variant, which for simplicity we refer to as VRNN. In practice, these two architectures generally demonstrate similar levels of performance  \cite{chung_recurrent_2016,gedon_deep_2021}.

\subsection{Ensemble Analysis}
The probabilistic architectures described above help to address the uncertainty due to limited training data. However, there is also uncertainty due to the random nature of the optimization algorithm used to train the network and the highly non-convex nature of the optimization landscape. To leverage this uncertainty we employ an ensemble approach in which we train the same architecture multiple times on the same training data. This results in an ensemble of NN's: $\mathcal{G}_{\theta_j}$ and therefore an ensemble of predictions $\hat{\mathbf{u}}_j = \mathcal{G}_{\theta_j}[\mathbf{v}], ~\ j=1...N_{e}$ where $N_{e}$ is the number of ensemble members. We then define the prediction of any statistic or observable $g(\mathbf{u})$ as the average prediction of the ensemble members 
\begin{equation}\label{eq:mean_pred}
    \bar{g} = \frac{1}{N_{e}}\sum_{j=1}^{N_{e}} g(\mathcal{G}_{\theta_j}[\mathbf{v}]).
\end{equation}
The uncertainty is then quantified through the ensemble variance
\begin{equation}\label{eq:var_pred}
    \sigma^2_g = \frac{1}{N_{e}-1}\sum_{j=1}^{N_{e}} \left(g(\mathcal{G}_{\theta_j}[\mathbf{v}])-\bar{g}\right)^2.
\end{equation}
We note that due to their probabilistic nature, each forward evaluation (on the same input) of the VAE-RNN, STORN, and VRNN architectures leads to slightly different outputs. However, we have found that the variance in the long time statistics of these variable predictions is negligible.
In fact, the variance quantified by \eqref{eq:var_pred} is dominated by the ensemble variance, and is not meaningfully affected by the probabilistic nature of the architectures. This is both expected and desirable, as even if each forward pass of the model produces a different realization, we expect each of these to be drawn from the same distribution and thus to share common long time statistics.

In \ref{app:model_selection} we present a detailed parametric study on the effects of ensemble size and training duration for each of the four architectures described above. In general, for all architectures the effect of considering an ensemble as opposed to a single network is small but meaningful. For clarity of exposition, we focus the remainder of our discussion on results computed from an ensemble of 6 neural networks each of which is trained for 500 epochs. We found that in general increasing the ensemble size further increased the computational cost substantially while leading to only marginal improvements. All following results -- for all architectures -- are the ensemble mean prediction as quantified by \eqref{eq:mean_pred}.

\subsection{Network Architecture and Training Details}
The correction operator used in this work are based on the LSTM-based architecture already validated by \citeA{barthel_sorensen_non-intrusive_2024} on the isotropic version of the QG model i.e. without topography. This architecture consists of a single layer encoder which compresses the input to a hidden state of dimension 60, followed by an LSTM layer of the same size, and a single layer decoder that restores the output to the original size. For the probabilistic models the latent space dimension was also set to $60$.

As our main aim in this paper is to exhibit the advantages of the probabilistic methods, we have left the encoder, decoder, and LSTM layers of the networks as unaltered as possible. With the exception of the VRNN architecture, the inclusion of the latent space does not meaningfully impact the number of trainable parameters which are summarized in table \ref{tab:ml_dof}.  The increase in degrees of freedom for the VRNN architecture is due to the increased size of the input to the decoder layer \eqref{eq:vrnn_decoder}.
However, we found that neither increasing the depth or width of the encoder and decoder layers, nor varying the dimension of the latent space had any significant impact on the results. Therefore, we expect that any differences in performance are not simply due to an increase in the degrees of freedom. 

The loss function used to train the correction operators consists of three terms: a mean squared prediction error, a term that penalizes deviations in the conservation of a mass in the QG model, and the KL divergence term regularizing the latent space distribution -- the latter being only present for the probabilistic architectures.
The overall expression for the loss is given by
\begin{equation}
    L(\theta) = \int_0^T \left\|\mathcal{G}_{\theta}[\mathbf{v}_\tau']-\mathbf{u}\right\|^2dt  + \int_0^T \left\|\mathcal{G}_{\theta}[\mathbf{v}_\tau']\right\|dt  +  \lambda D_{KL}\left(\mathcal{N}\left(\boldsymbol{\mu}_z(\theta),\boldsymbol{\sigma}_z(\theta) \right),\mathcal{N}\left(\mathbf{0}, \mathbf{I}\right) \right)
\end{equation}
The normalization constant $\lambda$ sets the strength of the regularization on the probabilistic latent space: if it is too large, the model will ignore the prediction error and drive the latent space to pure white noise, and if it is too small, the model will over fit to the data and the latent space will have no effect. Empirically, we found that for our problem a value of $10^{-4}$ led to the best results.

\begin{table}[]
    \centering
    \begin{tabular}{c|c|c|c}
        Model & Trainable Parameters & $\overline{D_{KL}}$ & $\overline{L_{1}}$ \\
        \hline
        RNN & 168,492 & 0.0239 & 3.68\\ 
        VAE-RNN & 175,812 & 0.0026 & 4.35\\ 
        STORN & 190,212 & 0.0089 & 2.46\\ 
        VRNN & 259,332 & 0.0044 & 2.47
    \end{tabular}
    \caption{Number of trainable parameters (degrees of freedom) and global prediction errors for each network architecture considered in this work.}
    \label{tab:ml_dof}
\end{table}

\section{Results}\label{sec:Results}
Here we showcase the results of our machine learning framework, as introduced in \S\ref{sec:Math} using the network architectures described in \S\ref{sec:ML}, applied to the quasi-geostrophic system described in \S\ref{sec:QG}. All the results herein represent the ensemble mean prediction \eqref{eq:mean_pred} of six ML correction operators applied to a single unseen realization of the flow of length 34,000 time units -- 34 times the length of the training data. The focus of the discussion is the comparison of the architectures described in \S\ref{sec:ML}; ensemble size sensitivity is explored in \ref{app:model_selection}.

We present our results in the form of probability density functions (pdfs) as well as one and two point correlations. We are interested in the ability of the correction operator to accurately quantify the probability of extreme events -- particularly of those whose return period is longer than the training data. Therefore, all pdf results will be presented on both a linear and logarithmic scale.
The former illustrates the bulk of the distribution, while the latter emphasizes the tails.
Accordingly, we will make use of the following two error metrics to evaluate the statistical accuracy of the ML predictions. The Kullback-Liebler (KL) divergence, defined as
\begin{equation}\label{eq:KLD}
    D_{KL}(p||q) \equiv \int p(x)\log\left(\frac{p(x)}{q(x)}\right)dx 
\end{equation}
and the L1 error of the log-pdf
\begin{equation}\label{eq:L1}
    L_{1}(p||q) \equiv \int |\log\left(p(x)\right) - \log\left(q(x)\right)|dx.
\end{equation}
This latter metric, which we will refer to as the $L_1$ error, is chosen specifically to emphasize deviations in the tails. These two metrics can be thought of as measures of overall and extreme event specific accuracy respectively.

\subsection{Global Statistics}
\begin{figure}
    \centering
    \includegraphics[width=0.95\textwidth]{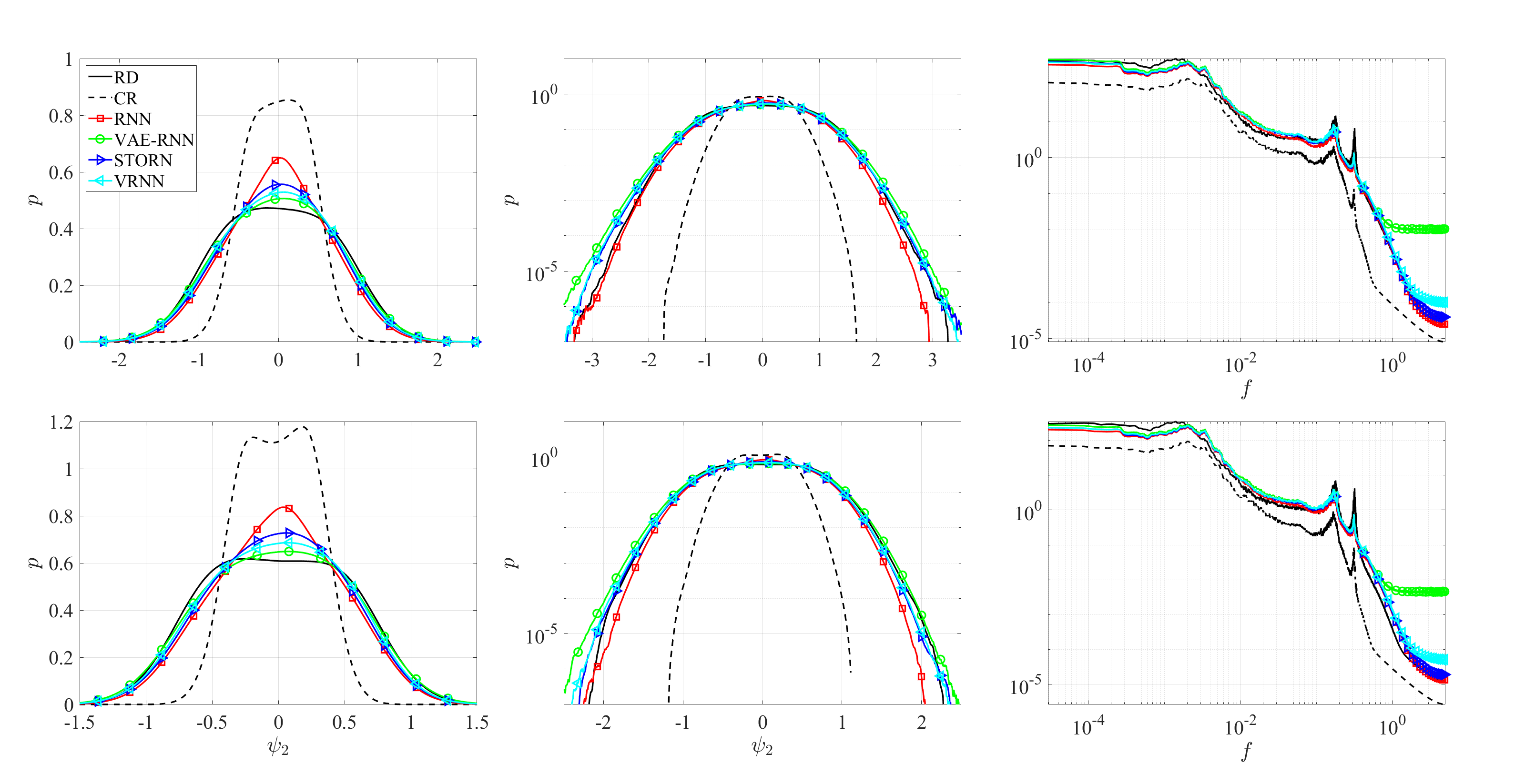}
    \caption{Global pdf, log-pdf, and PSD of $\psi_1$ (upper panel) and $\psi_2$ (lower panel). RD (solid black), CR (dashed black), RNN (red), VAE-RNN (green), STORN (blue), VRNN (teal).}
    \label{fig:global_stats}
\end{figure}

Results for the global pdf, log-pdf, and power spectral density of the stream function are shown in Figure \ref{fig:global_stats}. Here we compare the (ensemble mean) prediction of the ML corrected coarse model (shown in color) to the true statistics (solid black) and those of the uncorrected coarse model (dashed black). All three probabilistic architectures capture the true pdf better than the deterministic architecture -- which while significantly improving the uncorrected simulation, still overestimates the probability of very low amplitude events and under estimates the tail statistics. The average (over $\psi_1$ and $\psi_2$) global KL-Divergence and L1 log-pdf error for each architecture is listed in table \ref{tab:ml_dof}. In all cases, the probabilistic architectures outperform the deterministic RNN. The VAE-RNN achieves the lowest overall KL divergence, but has the highest $L_1$ error, meaning it captures the bulk of the distribution well but does not capture the tails accurately. In regard to capturing tail risk events, the STORN and VRNN architectures generally provide optimal results. They accurately reflect the true distribution across the full range of amplitudes, while the VAE-RNN architecture tends to mildly over-predict the tails.

\begin{figure}
    \centering
     \includegraphics[width=0.95\textwidth]{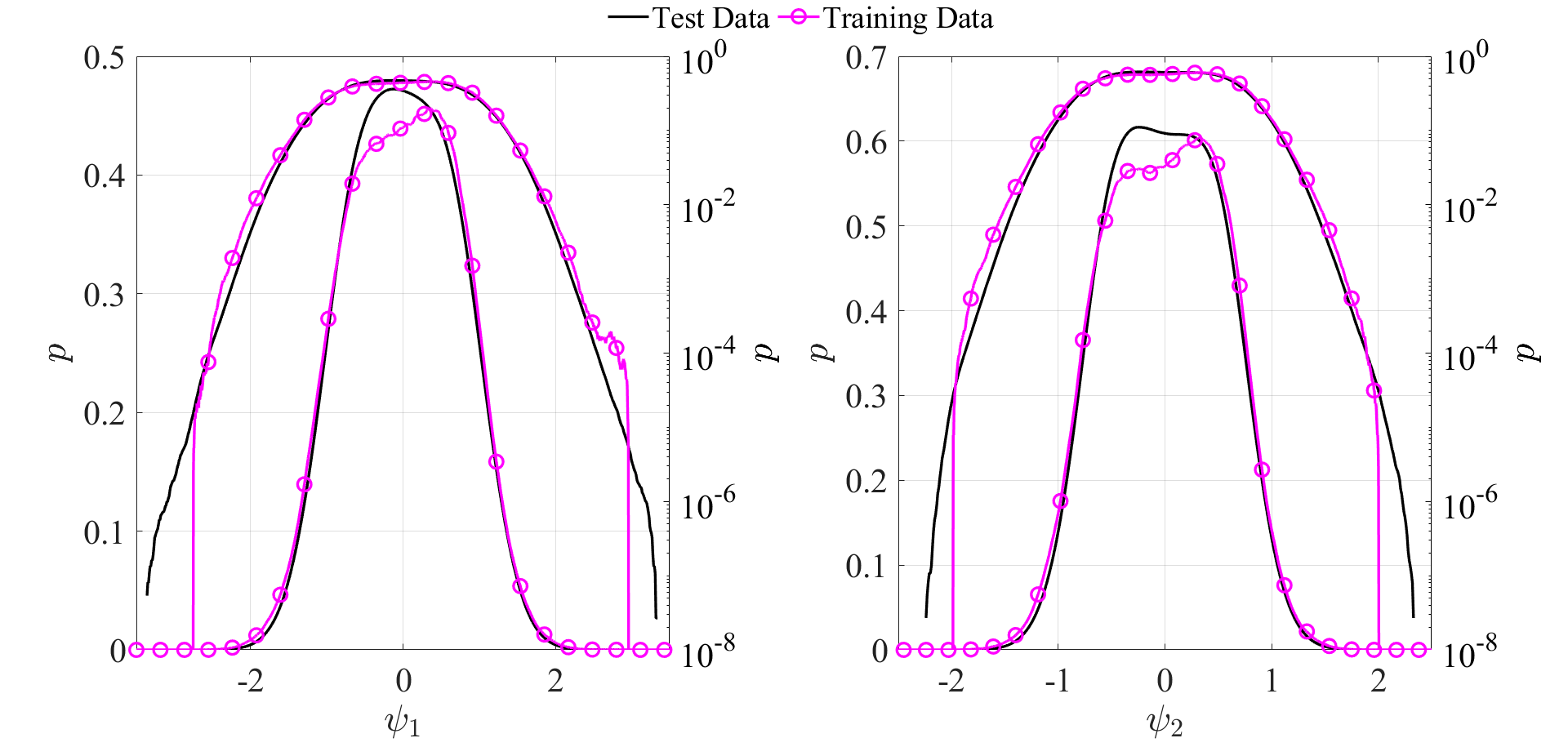}
         ~\ ~\  ~\ ~\ ~\ ~\ ~\ ~\ ~\ ~\ ~\ ~\ ~\ ~\ ~\ (a) ~\ ~\ ~\ ~\  ~\ ~\ ~\ ~\ ~\ ~\ ~\ ~\ ~\ ~\ ~\ ~\ ~\  ~\ ~\ ~\ ~\  ~\ ~\ ~\  ~\ ~\ ~\ (b)   
    \caption{Ground truth reference statistics of 34,000 time unit test data (black) and 1,000 time unit training data (magenta) of $\psi_1$ (a) and $\psi_2$ (b). Each subfigure shows the pdf on linear and logarithmic scale.  }
    \label{fig:training_data}
\end{figure}

To highlight the ability of our ML correction operator to extrapolate from the short training data we show in figure \ref{fig:training_data} the differences in the statistics of the long (34,000 time unit) test data and the short (1,000 time unit) training data. The training data is clearly not converged. In fact, the heavy tails are missing from the training data entirely. As shown in figure \ref{fig:global_stats}, the ML corrections accurately capture the tails of the underlying pdf even where there the training data does not. From this ability of the ML correction to extrapolate beyond the training data we infer that the NN is in fact learning some notion of the underlying system dynamics -- a key feature in extending the proposed method to more complex system and even longer time horizons.

In figure \ref{fig:global_stats} we also plot the global power spectral density (PSD), defined as the spatial average of the temporal Fourier transform of the autocorrelation,
\begin{equation}
    S_j\left(f\right) \equiv \int_0^{2\pi} \int_0^{2\pi} \int R_j(\tau) e^{-if\tau} d\tau dx dy,
\end{equation}
\begin{equation}
    R_j\left(\tau \right) \equiv \int \psi_j(t)\psi_j(t+\tau)dt.
\end{equation}
With the exception of the VAE-RNN architecture, the ML corrections accurately reflect the true power spectrum across the full range of frequencies -- including the two characteristic peaks near $f = 0.1$. The VAE-RNN architecture accurately captures the lower frequencies -- those with meaningful energy content -- but fails to accurately predict the energy roll off of the highest frequencies. This is an intrinsic limitation of the VAE-RNN architecture \eqref{eq:rnn_vae_rnn}. For frequencies with very low energy the prediction error term in the loss function will become negligibly small, and the training loss will be dominated by the term enforcing the white noise prior placed on the latent space. For those frequencies, the latent space $z$ will then be driven to exactly white noise, and due to the lack of communication across the time steps of $z_t$ inherent in the solely downstream latent space interaction in \eqref{eq:rnn_vae_rnn} the output will also be dominated by white noise. This flat spectrum phenomenon is also present to a minor extent in the VRNN architecture (Fig. \ref{fig:global_stats}) which also has a downstream latent space dependency. However, the inclusion of the upstream dependency in the VRNN architecture enables the communication between time steps $z_t$ which helps to additionally regularize the latent space. Finally, we again emphasize the extrapolation capabilities of our training framework evidenced by the difference between the PSD of the training data (magenta) and the test data (black).


\begin{figure}
    \centering
    \includegraphics[width=0.95\textwidth]{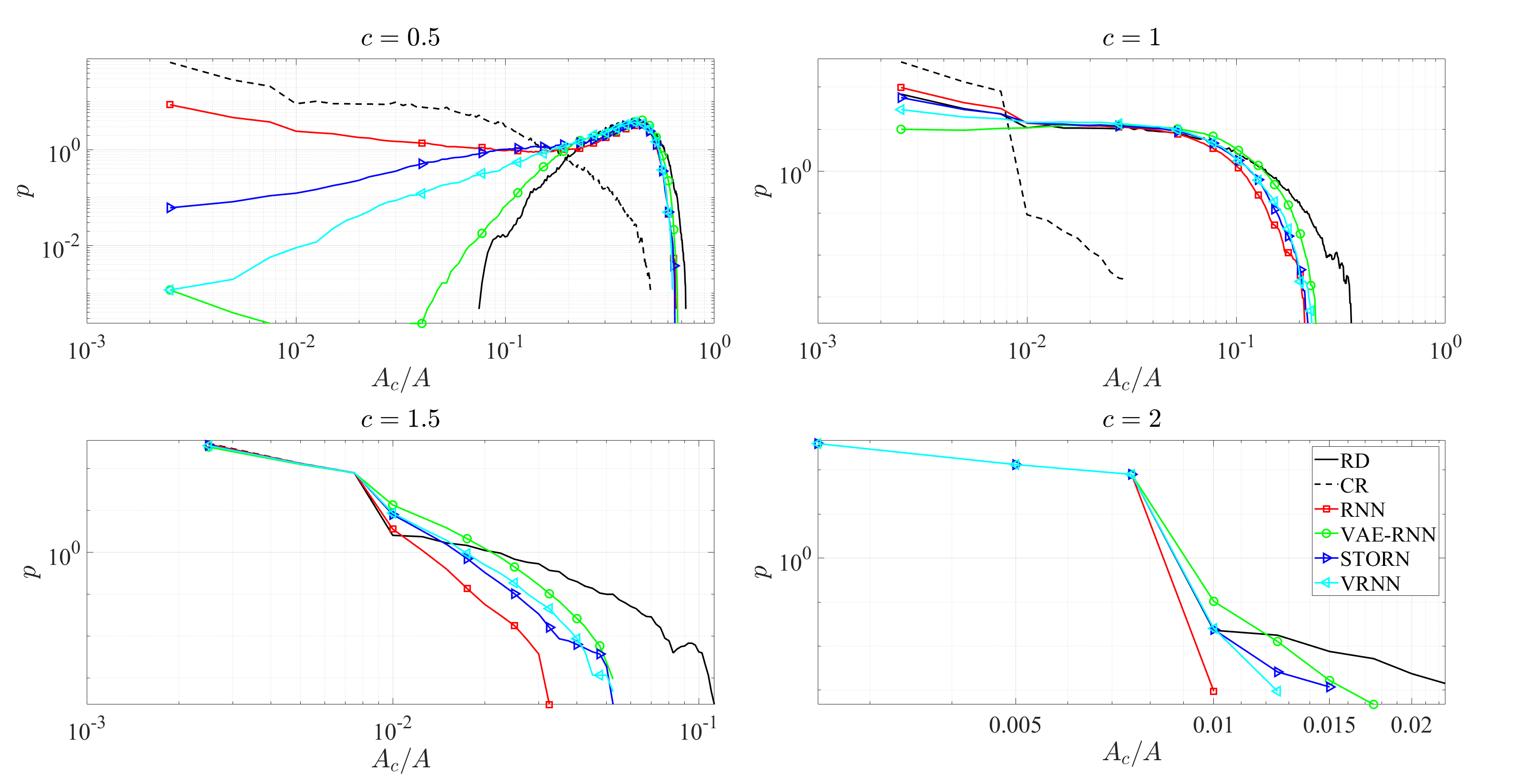}
    \caption{Pdf of fraction of domain over which $|\psi_2|$ exceeds fixed threshold $c$ for range of $c \in [0,2]$. RD (solid black), RNN (red), VAE-RNN (green), STORN (blue), VRNN (teal).}
    \label{fig:aot_psi2}
\end{figure}

To further probe the spatiotemporal accuracy of the ML corrected fields we compute the fraction of the domain over which the stream function exceeds a certain threshold as a function of time,
\begin{equation}
    A_c(t)/A = \frac{1}{N_x N_y} \sum_{i,j}^{N_x, N_y} H\left( |\psi(x_i,y_j,t)| - c\right).
\end{equation}
Here $c$ is the given threshold and $H(x)$ is the unit step function such that $H(x) = 1$ if $x\ge0$ and $H(x) =0$ if $x<0$. This metric characterizes how reliably the ML corrections can capture the frequency and spatial extent of extremes, and is a proxy for the ability of the model to capture large-scale extreme phenomena in climate models, such as heatwaves. The probability density functions of $A_c(t)/A$ for a range of $c$ are plotted in Figure \ref{fig:aot_psi2}. For brevity we focus on $\psi_2$; results for $\psi_1$ are included in \ref{app:psi_1_results}. First, we note that the uncorrected (CR) solution vastly underestimates the amplitude of the true solution -- missing the higher-amplitude extremes entirely. In contrast, all ML correction models are able to capture the bulk of the distribution. Compared to the RNN, the probabilistic architectures track the pdf significantly better, with the VAE-RNN demonstrating the best performance. The deterministic RNN on the other hand significantly overestimates the probability of low area ratios for the lower thresholds $c<1$. This is consistent with the results in Figure \ref{fig:global_stats}, where the deterministic RNN significantly overestimates the likelihood of very low amplitudes. The probabilistic architectures also seem to demonstrate marginal improvements for higher values of $c$. However, in these cases the sample size is small and the pdfs -- computed by Monte Carlo sampling -- are clearly not fully converged.

\subsection{Regional Statistics}
Due to the anisotropic nature of the QG flow under consideration we are particularly interested in the regional variation of the quality of the ML correction. Therefore, in addition to the global statistics, we also analyze the statistics as a function of spatial location. For clarity of exposition we will focus here on the results in the lower layer, $\psi_2$. The corresponding results for the upper layer, $\psi_1$ -- which are qualitatively similar -- are summarized in \ref{app:psi_1_results}.

\subsubsection{Single-Point Statistics}
\begin{figure}
        \centering
        \begin{tabular}{ll}
       \includegraphics[width=0.95\textwidth]{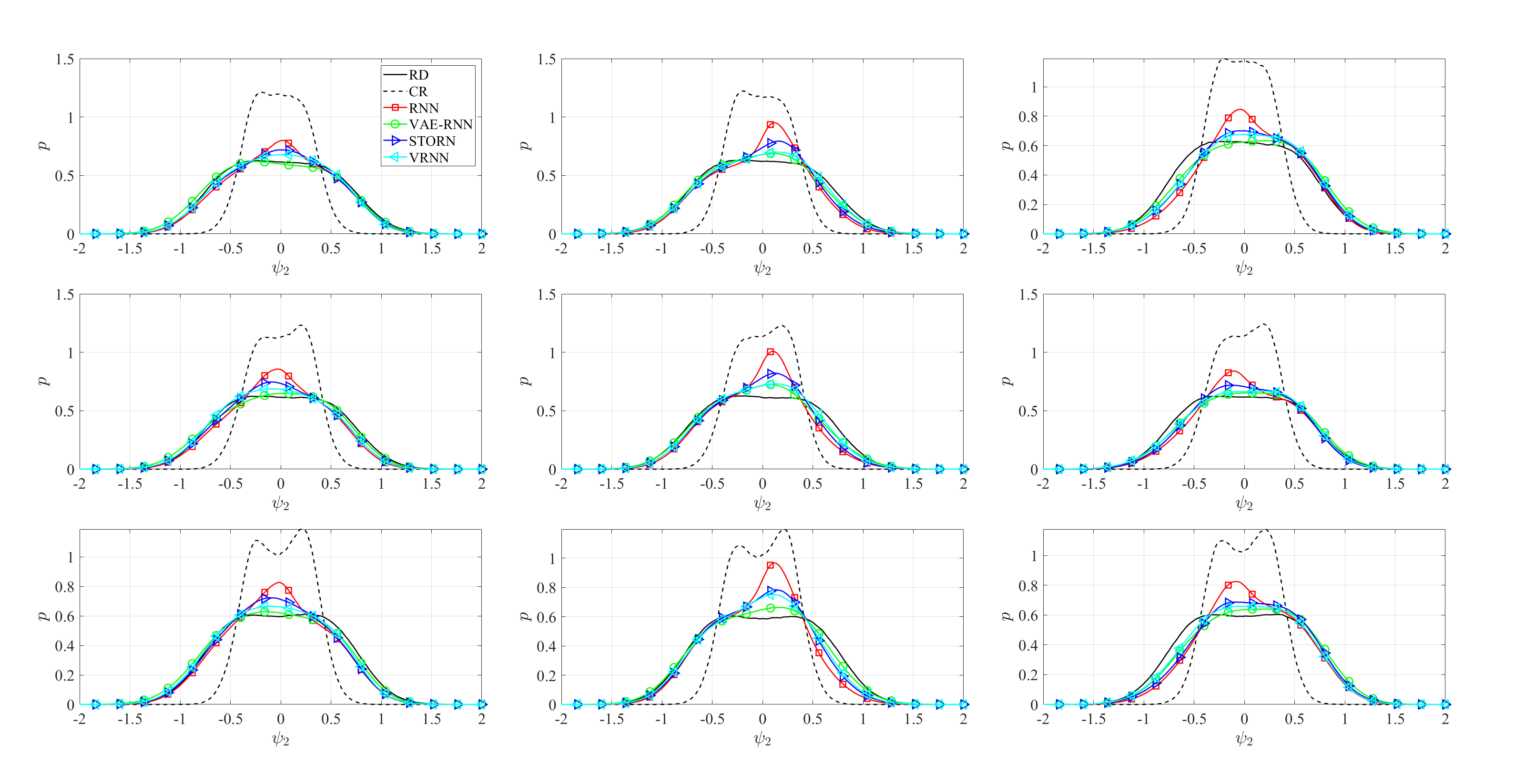}
      \\
    ~\  ~\    ~\  ~\    ~\   ~\   ~\   ~\   ~\   ~\  ~\   ~\  ~\  ~\   ~\   ~\  ~\  ~\  ~\   ~\   ~\   ~\  ~\  ~\  ~\  ~\  ~\  ~\  ~\  ~\  ~\  ~\  (a)  \\
        
        \includegraphics[width=0.95\textwidth]{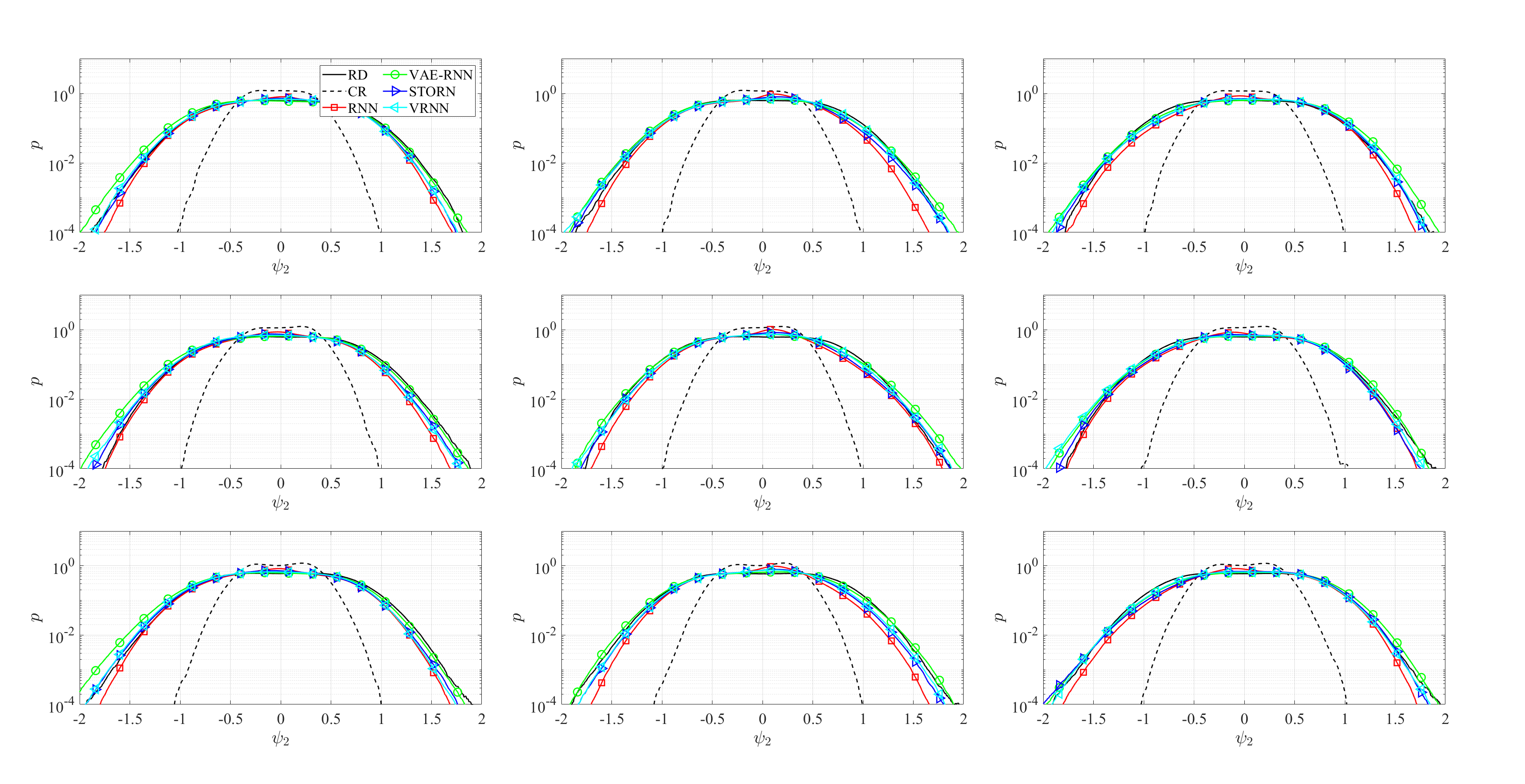}
        \\
          ~\  ~\    ~\  ~\    ~\   ~\   ~\   ~\   ~\   ~\  ~\   ~\  ~\  ~\   ~\   ~\  ~\  ~\  ~\   ~\   ~\   ~\  ~\  ~\  ~\  ~\  ~\  ~\  ~\  ~\  ~\  ~\  (b)  \\
        
        \end{tabular}
        \caption{Regional pdf (a) and log-pdf (b) of $\psi_2$.  RD (solid black), CR (dashed black), RNN (red), VAE-RNN (green), STORN (blue), VRNN (teal). }
        \label{fig:regional_pdf}
\end{figure}

We first illustrate our results in terms of single point statistics in the form of the pdf and log pdf. The regional power spectra show very little regional variation so we omit them here. We divide the domain $[x,y] \in [0,2\pi]$ into a $3\times3$ grid 
and compute the statistics of the stream function in each sub-region. Figure \ref{fig:regional_pdf}a and b show the pdf and log-pdf of the $\psi_2$ in each sub-region. The difference in pdf shape with respect to location is seen most clearly in the asymmetry of the uncorrected coarse pdfs -- some are clearly bimodal, while some peak at small negative values and others peak at small positive values.
As was the case with the global statistics, the probabilistic architectures demonstrate a clear improvement over the RNN in the ability to correct the local pdfs. Specifically, the latter incorrectly predicts peaks in the pdf near $\psi_2=0$ -- a feature which is significantly ameliorated by the probabilistic models, particularly the VAE-RNN and VRNN architectures. In many cases, the overpredictions by the RNN seem to be correlated with the previously mentioned anisotropic peaks in the pdfs of the uncorrected coarse data. This suggests an increased ability of the probabilistic models to handle anisotropic data. This is perhaps due to their ability to more efficiently encode complex (anisotropic) features which had not been seen in training.

\begin{figure}
    \centering
    \includegraphics[width=0.95\textwidth]{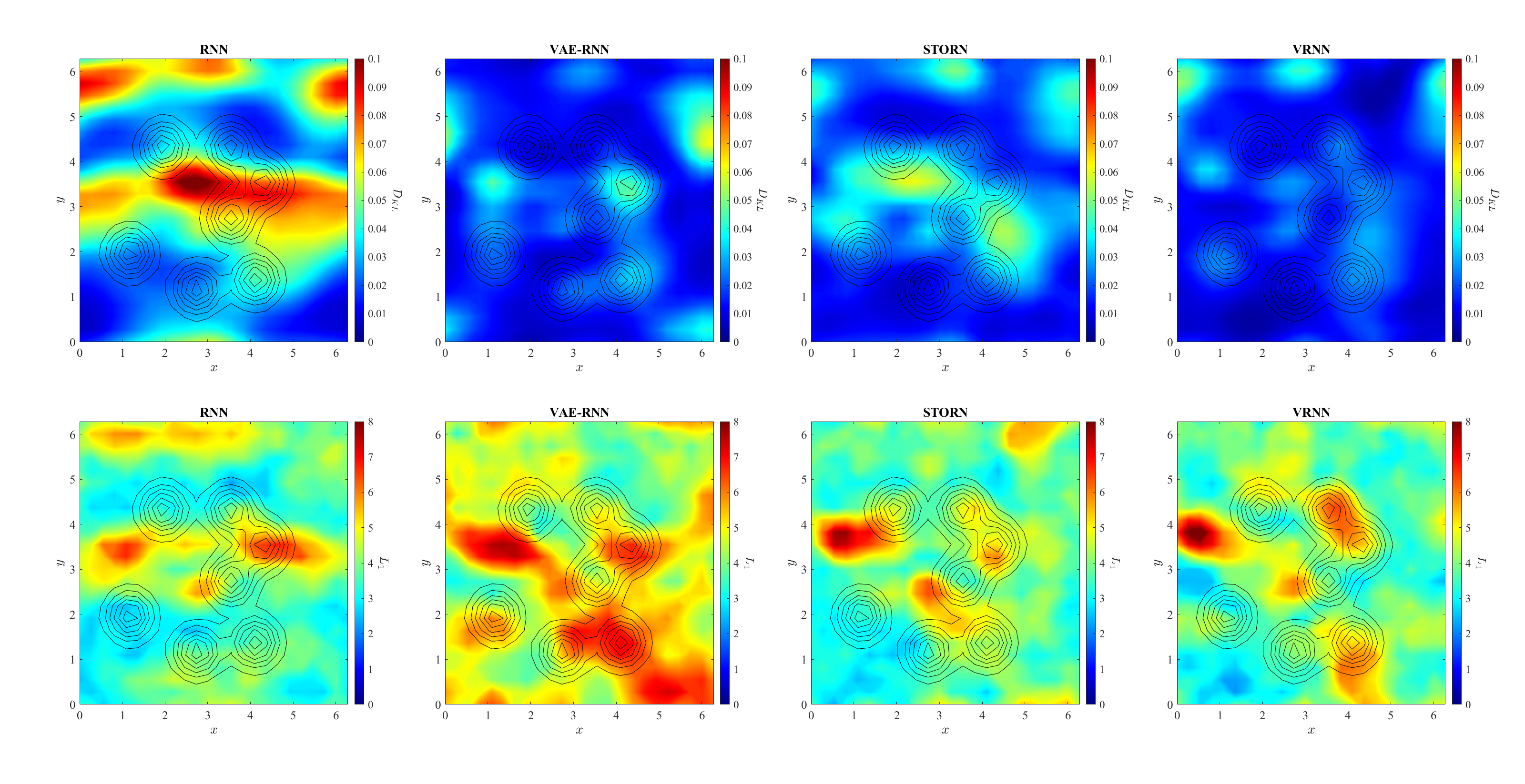}
    \caption{Spatial distribution of KL divergence (upper panel) and $L_1$ metric (lower panel) for $\psi_2$. From left to right: RNN, VAE-RNN, STORN, VRNN. The topography profile is show in black.}
    \label{fig:regional_pdf_error}
\end{figure}

A more quantitative view of the regional distribution of the ML correction is given in Figure \ref{fig:regional_pdf_error}, where the KL divergence and $L_1$ error are shown as a function of $x$ and $y$ coordinates -- here the pdfs and error metrics are computed at each grid point individually. All three probabilistic architectures outperform the deterministic RNN in terms of KL divergence relative to the true pdf. The VAE-RNN architecture has the highest $L_1$ error, while the RNN, STORN, and VRNN models show similar performance. As a reference we also plot the topography in solid black contours, and we note that the errors in the ML prediction are generally clustered immediately upstream of the topography profile. This is possibly due to an increase in complexity of the flow in this region. It is more likely that the ML correction operator will encounter vortical structures in testing that were not observed in the short training data set which may lead to higher errors.

\subsubsection{Fourier Cross-Correlations}
\begin{figure}
    \centering
    \includegraphics[width=0.95\textwidth]{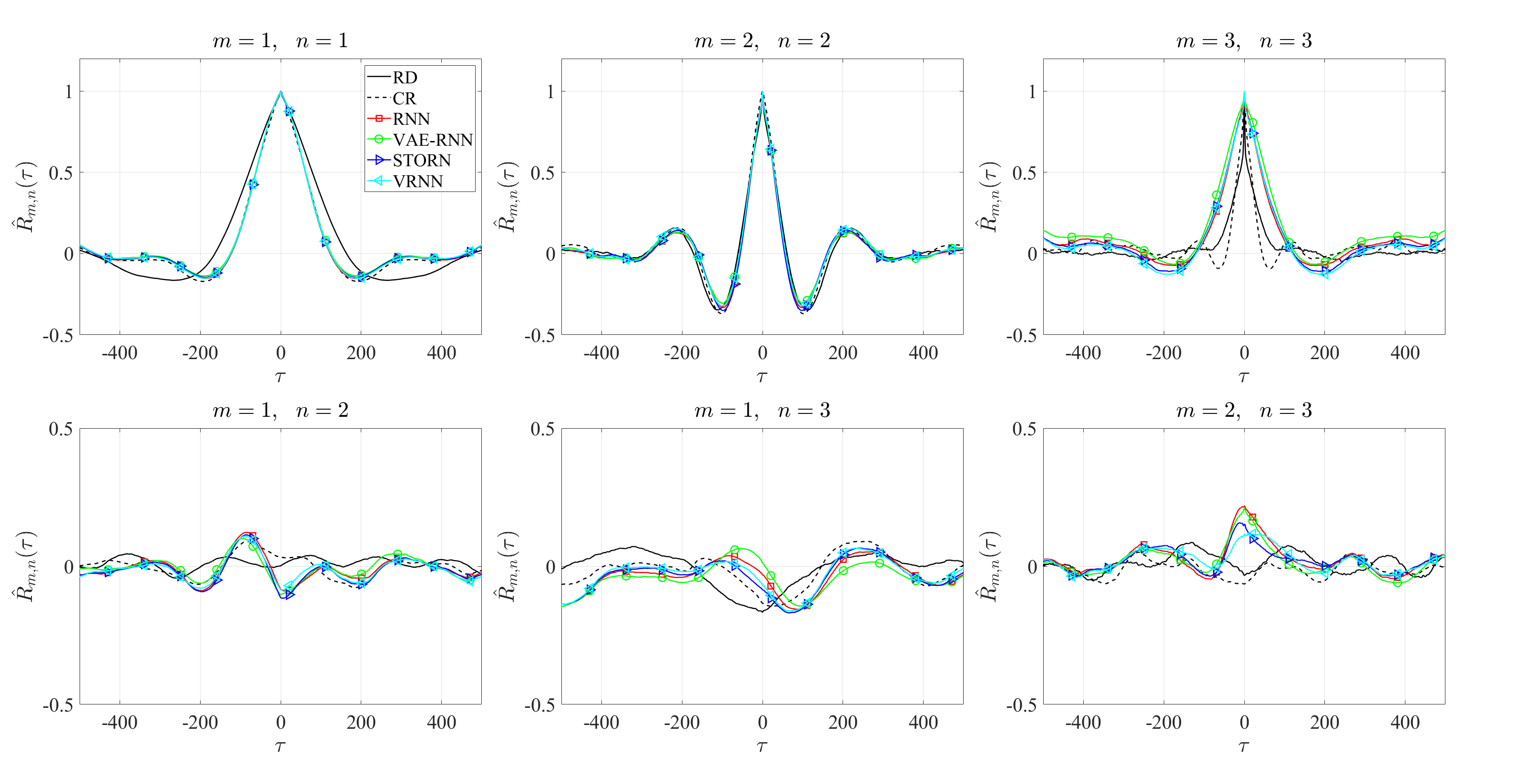}
    \caption{Normalized correlation between three largest zonally constant Fourier modes of $\psi_2$. RD (solid black), CR (dashed black), RNN (red), VAE-RNN (green), STORN (blue), VRNN (teal).}
    \label{fig:xcorr}
\end{figure}

To further investigate the spatiotemporal statistics of the corrected fields we compute the normalized cross-correlation between individual Fourier modes
\begin{equation}
    \hat{R}_{j,m,n} \equiv \frac{\int\hat{\psi}_{j}(\textbf{k}_m,t)\hat{\psi}_{j}(\textbf{k}_n,t+\tau)dt}{\sqrt{\int\hat{\psi}^2_{j}(\textbf{k}_m,t)dt\int\hat{\psi}^2_{j}(\textbf{k}_n,t)dt}}.
\end{equation}
We focus our discussion on the zonally constant modes, with wave number $\mathbf{k}_m = [0,k_m]$. If $m=n$, this metric is equivalent to a normalized autocorrelation, and for the case $m\neq n$ this metric can be interpreted as a phase shift between Fourier modes. The results for the three largest modes are shown in Figure \ref{fig:xcorr}. We find that the uncorrected coarse model correlations are already very similar to those of the high resolution reference. Therefore, the effects of the ML correction on this metric are marginal. In all cases we observe similar decorrelation profiles (top row of Fig. \ref{fig:xcorr}) -- with the ML correction affording a marginal improvement over the uncorrected baseline. The cross-correlations between Fourier modes (top row of Fig. \ref{fig:xcorr}) all fluctuate near 0 for all $\tau$, but again for all architectures we see minimal affect of the ML correction. 
One potential strategy to address this shortfall in the future is through network architectures which operate directly in Fourier space \cite{li_fourier_2021} -- an approach which has been demonstrated to be effective in modeling turbulent flows including global weather patterns \cite{li_fourier_2021,pathak_fourcastnet_2022} .

\FloatBarrier

\subsection{Spatiotemporal Features}
\begin{figure}
    \centering
    \begin{tabular}{ll}
         \includegraphics[width=1\textwidth]{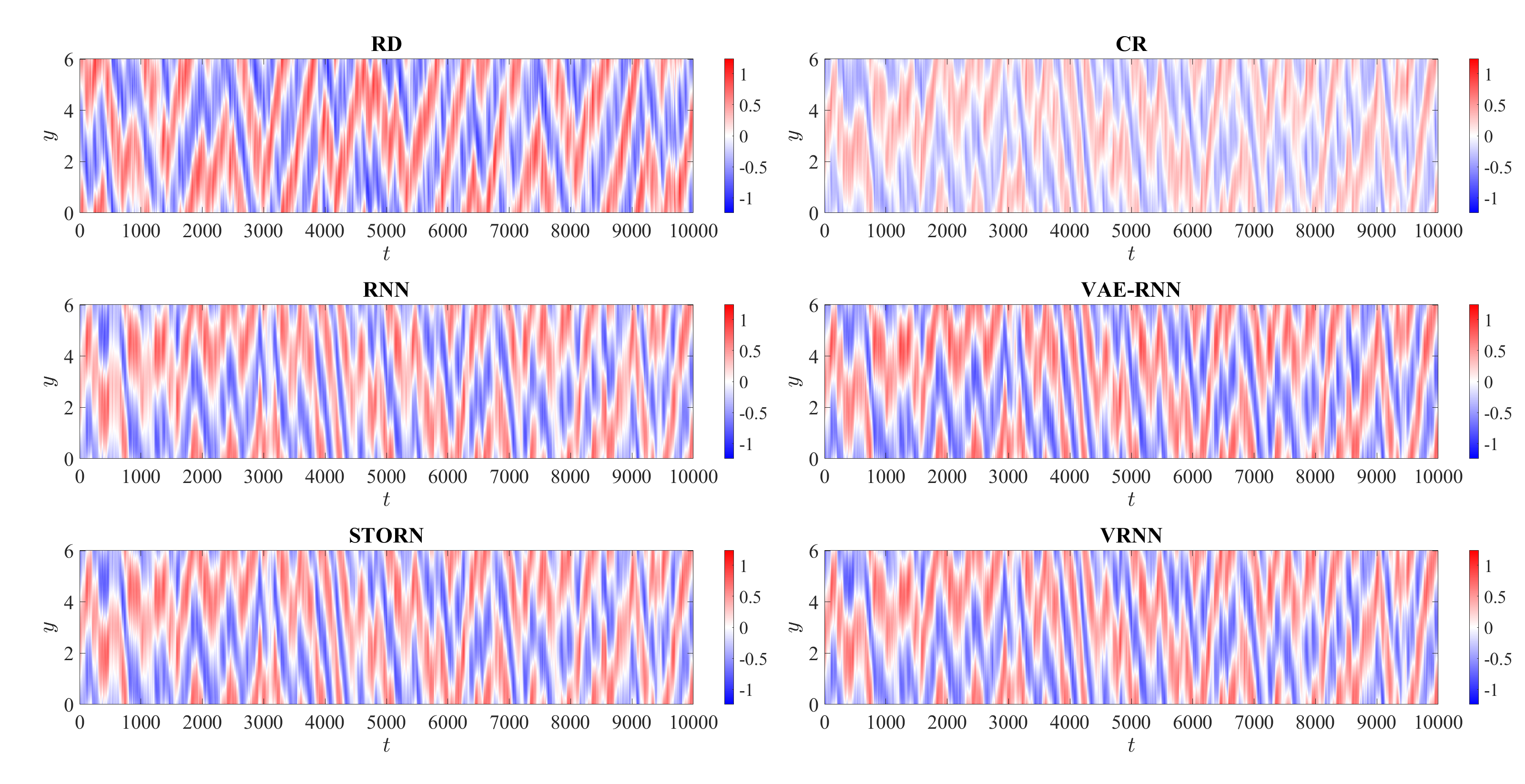} \\
    \end{tabular}
    
    \caption{Zonally averaged stream function $\bar{\psi}_2(y,t)$ . }
    \label{fig:zonal_mean}
\end{figure}

Finally, we investigate how the spatiotemporal features of the corrected flow fields compare to the reference solution. To this end we define the zonally averaged stream function
\begin{equation}\label{eq:psi_bar}
    \overline{\psi}_j(y,t) \equiv \frac{1}{2\pi}\int_0^{2\pi} \psi_j(x,y,t) dx,
\end{equation}
a quantity that enables us to analyze the meridional advection of structures in the field  \cite{hovmoller_1949, qi_using_2020}. Figure \ref{fig:zonal_mean} compares the zonally averaged flow field of the ML corrections to the RD and CR solutions. Since CR and RD are independent trajectories, we expect the corrected flow fields to share the statistics of the reference but not agree on a snapshot-by-snapshot basis. To improve the readability of the figure we limit the time axis to $10,000$ time units. The post-processed flow fields all display characteristic spatiotemporal structures which are consistent with the reference solution, and correct the significant magnitude underestimation of the coarse-resolution field.

\begin{figure}
    \centering
    \begin{tabular}{ll}
           \includegraphics[width=0.5\textwidth]{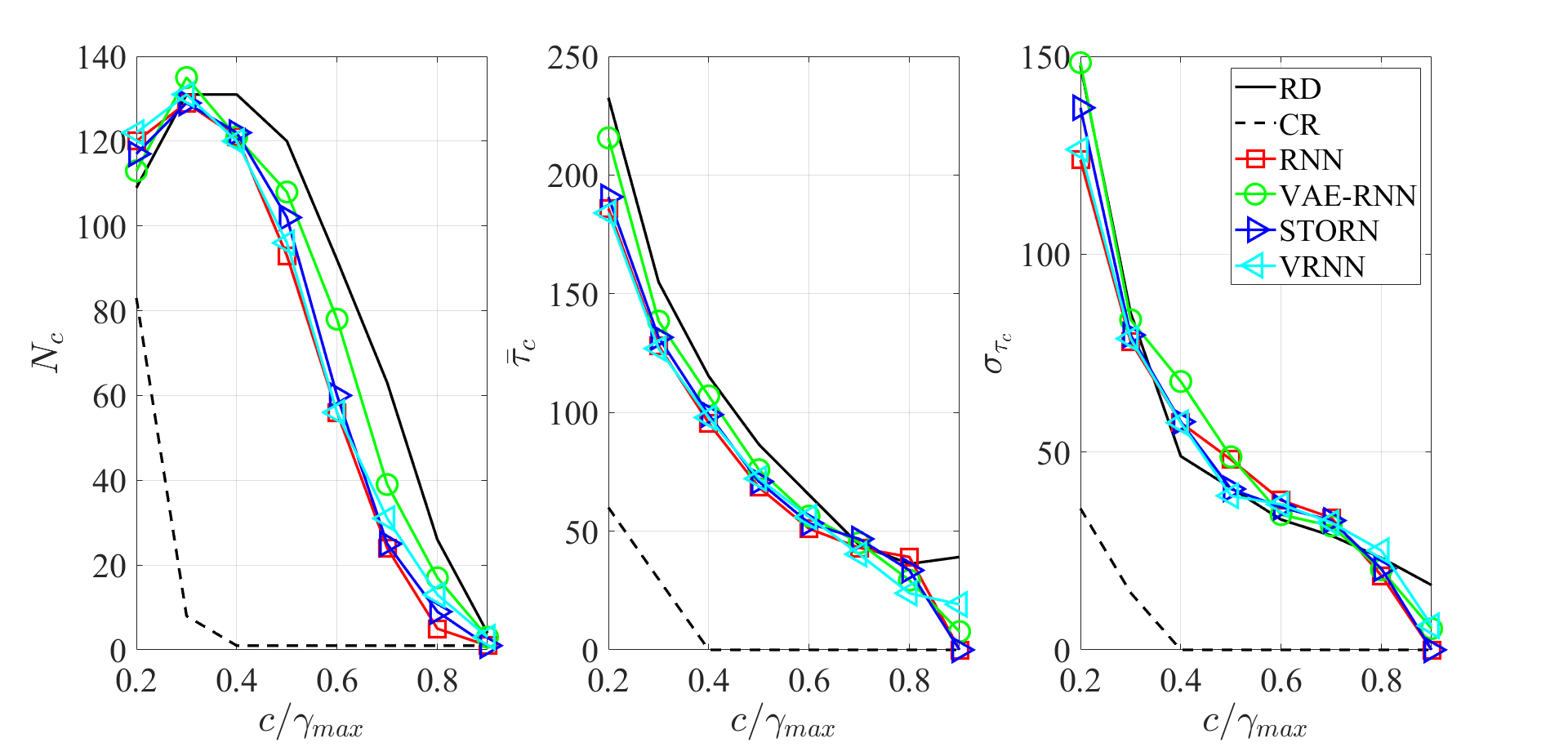}
           \includegraphics[width=0.5\textwidth]{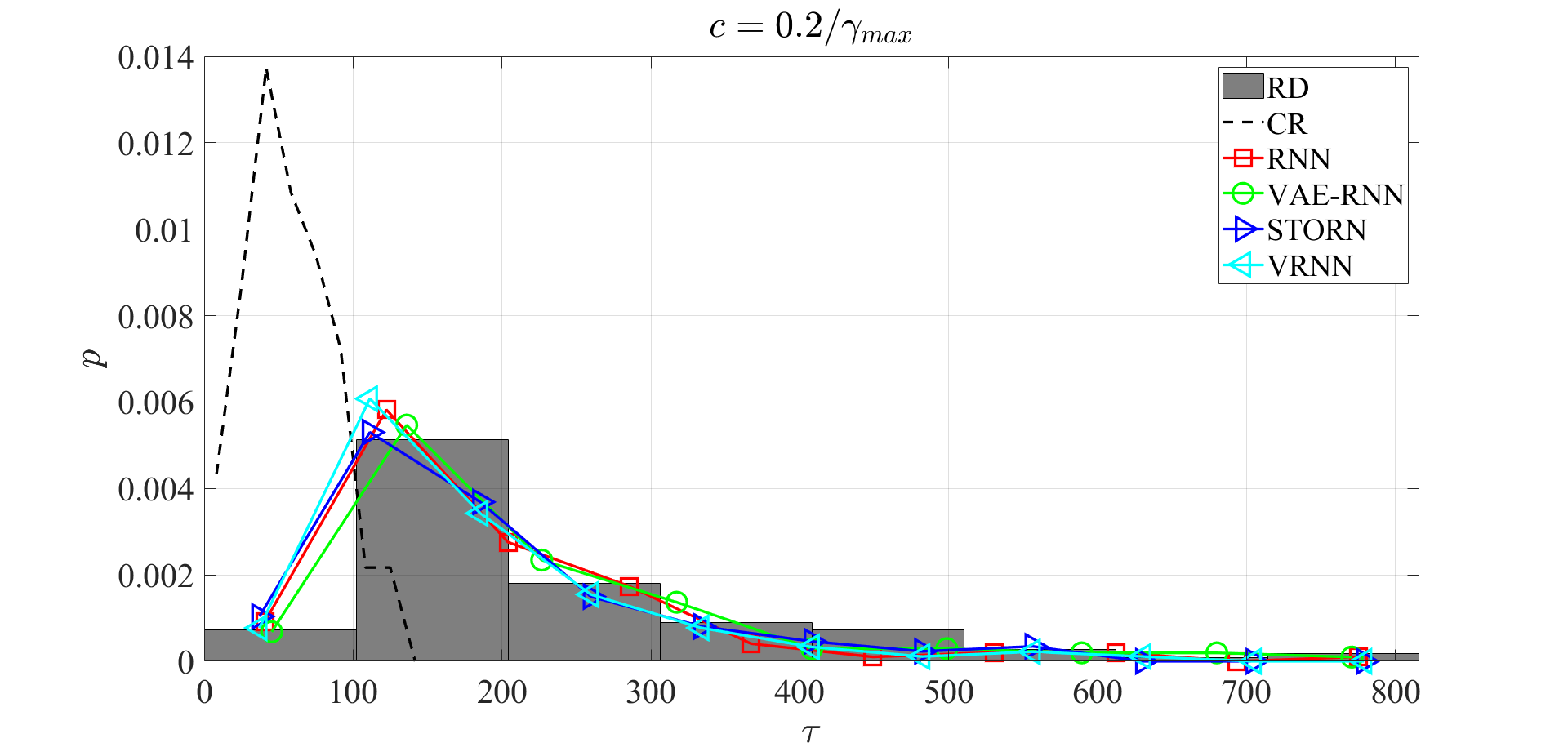} \\
            ~\ ~\  ~\ ~\ ~\ ~\ ~\ ~\ ~\ ~\ ~\ ~\ ~\ ~\ ~\ (a) ~\ ~\ ~\ ~\  ~\ ~\ ~\ ~\ ~\ ~\ ~\ ~\ ~\ ~\ ~\ ~\ ~\  ~\ ~\ ~\ ~\  ~\ ~\ ~\  ~\ ~\ ~\ (b)  \\
           \includegraphics[width=0.5\textwidth]{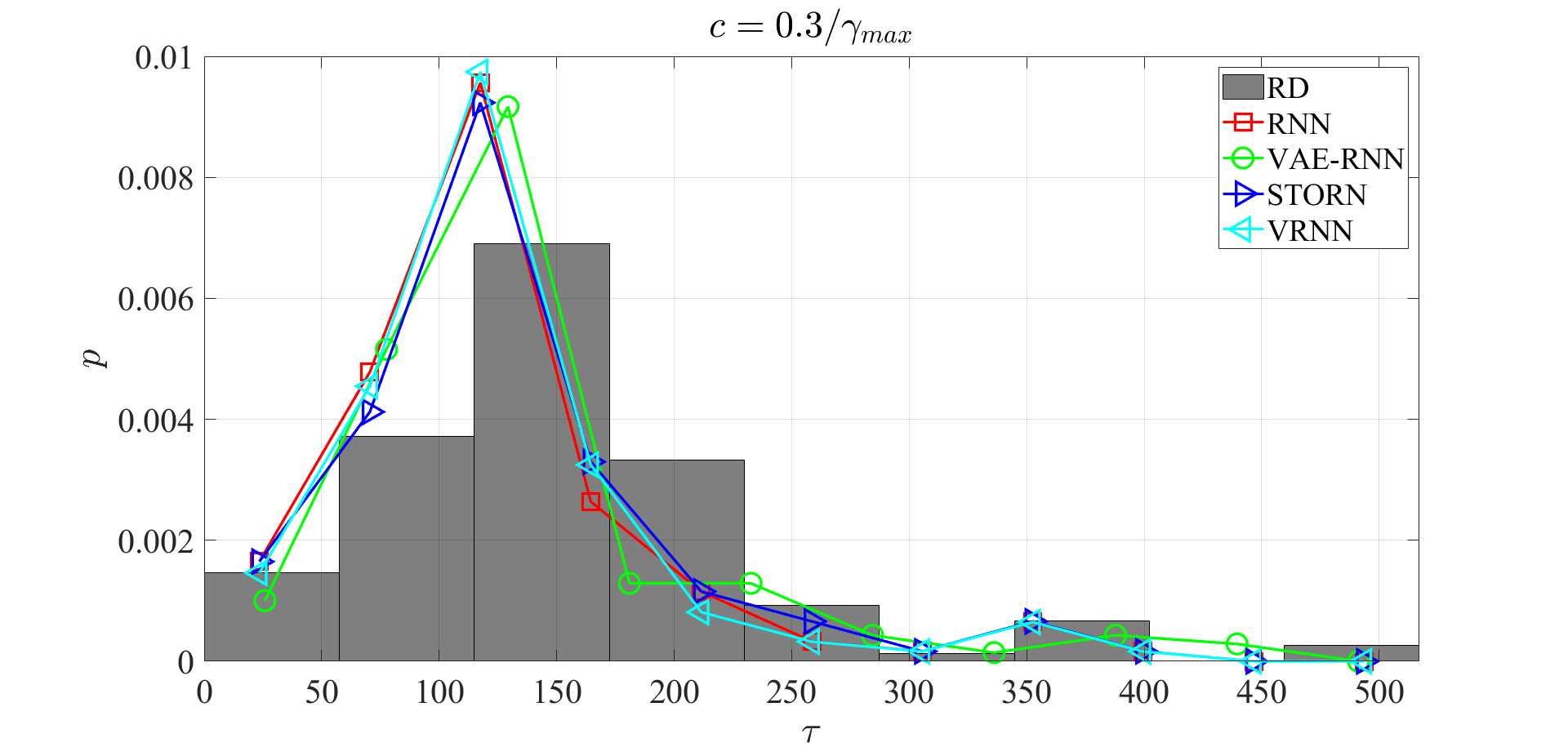} 
           \includegraphics[width=0.5\textwidth]{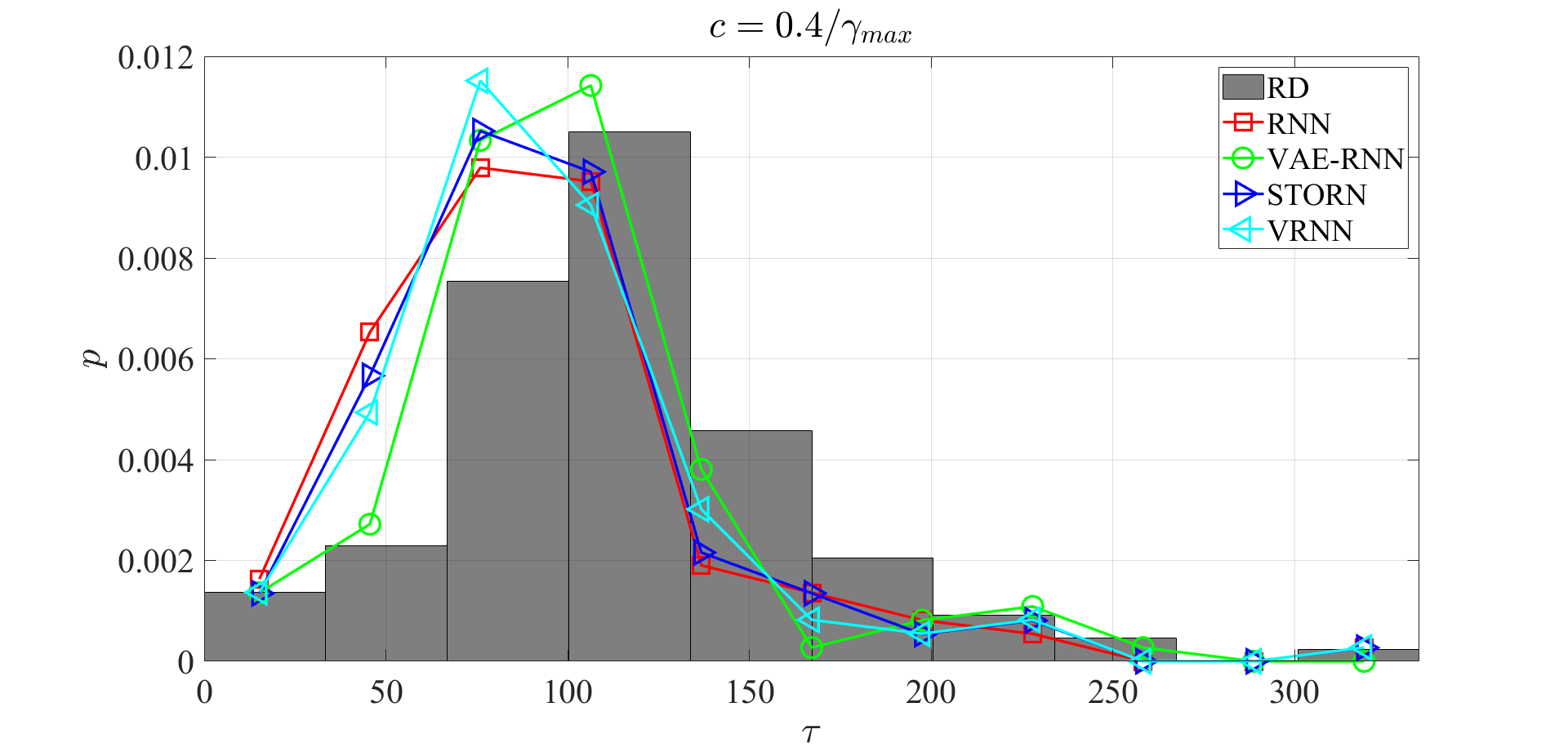} \\
             ~\ ~\  ~\ ~\ ~\ ~\ ~\ ~\ ~\ ~\ ~\ ~\ ~\ ~\ ~\ (c) ~\ ~\ ~\ ~\  ~\ ~\ ~\ ~\ ~\ ~\ ~\ ~\ ~\ ~\ ~\ ~\ ~\  ~\ ~\ ~\ ~\  ~\ ~\ ~\  ~\ ~\ ~\ (d)  \\
           \includegraphics[width=0.5\textwidth]{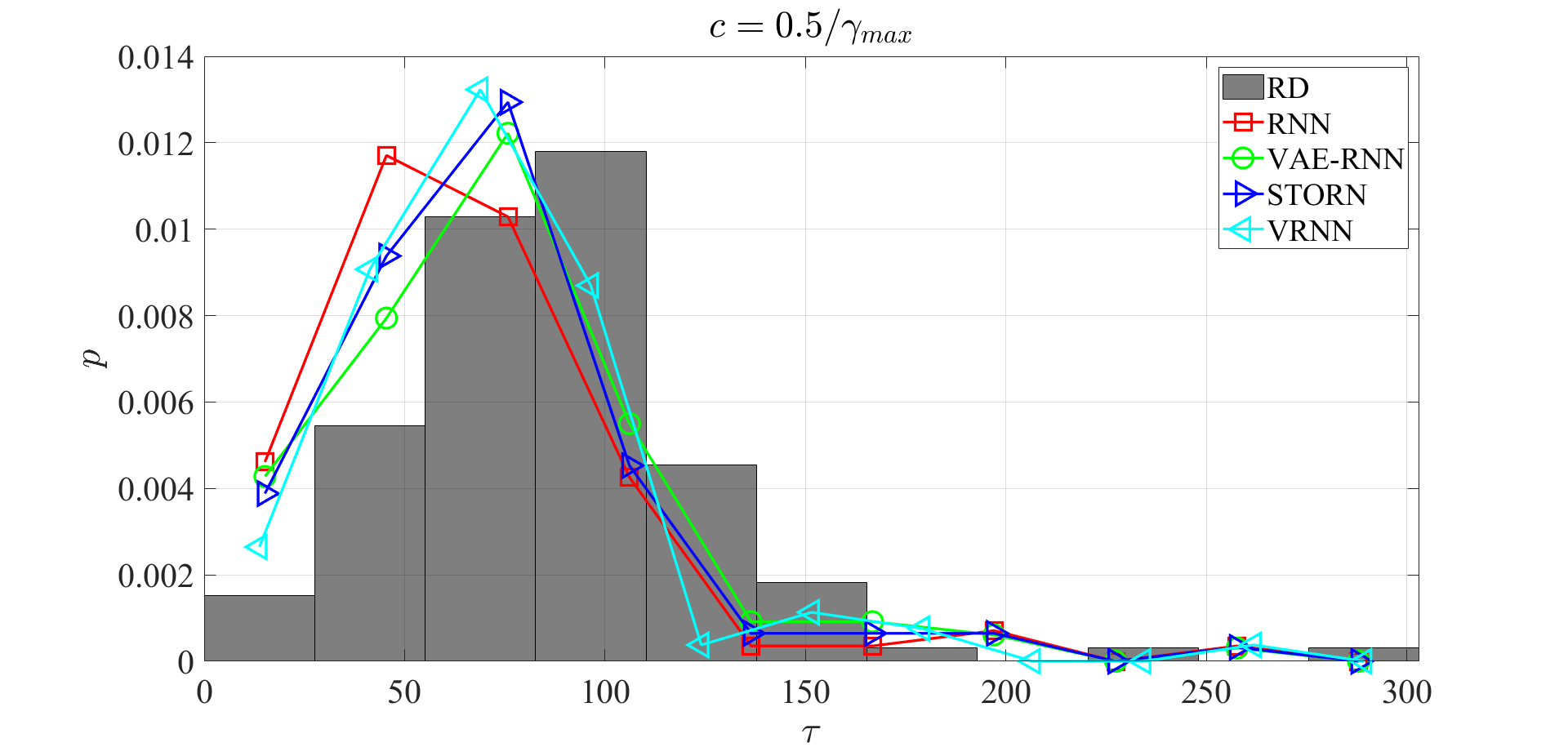} 
           \includegraphics[width=0.5\textwidth]{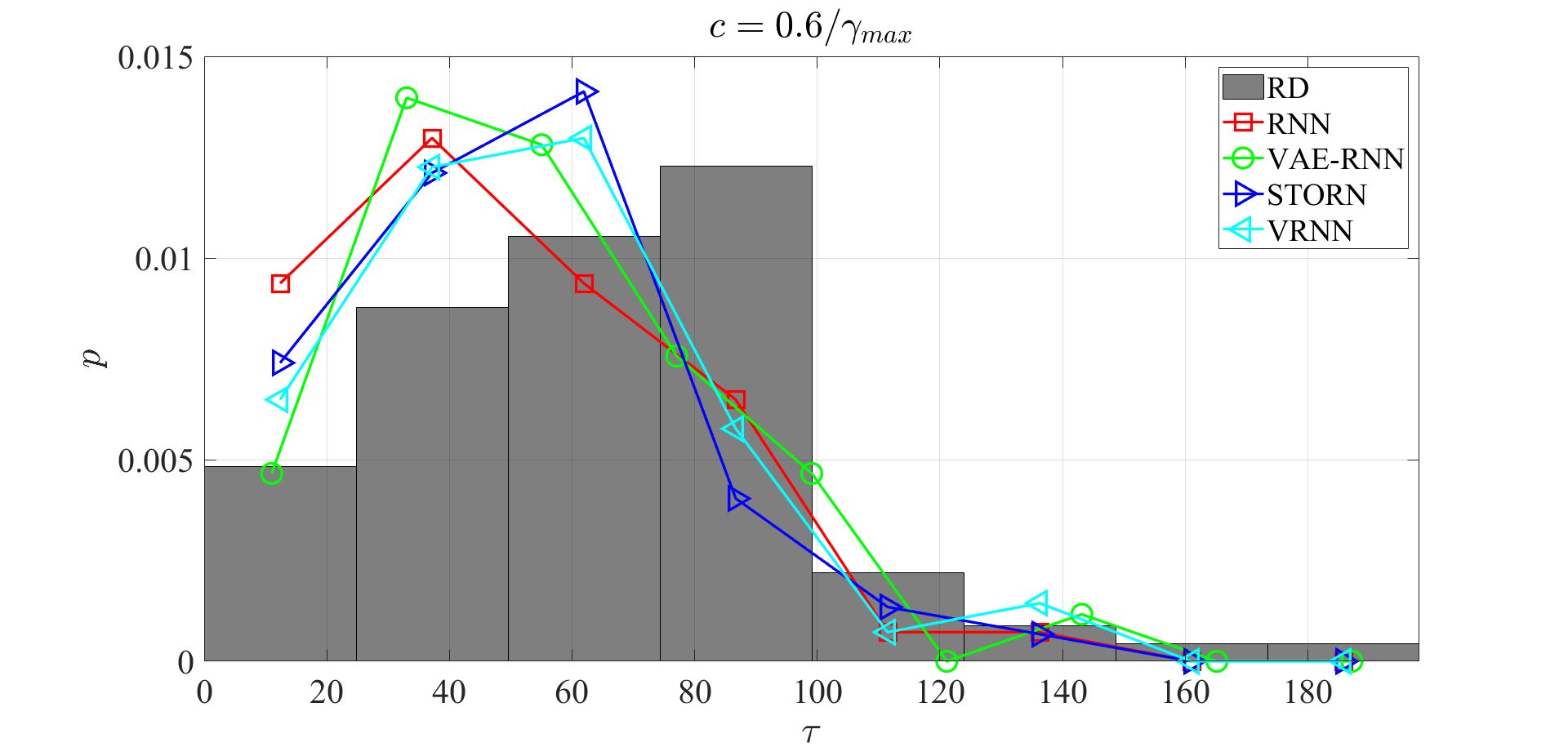} \\
            ~\ ~\  ~\ ~\ ~\ ~\ ~\ ~\ ~\ ~\ ~\ ~\ ~\ ~\ ~\ (e) ~\ ~\ ~\ ~\  ~\ ~\ ~\ ~\ ~\ ~\ ~\ ~\ ~\ ~\ ~\ ~\ ~\  ~\ ~\ ~\ ~\  ~\ ~\ ~\  ~\ ~\ ~\ (f)  \\
           \includegraphics[width=0.5\textwidth]{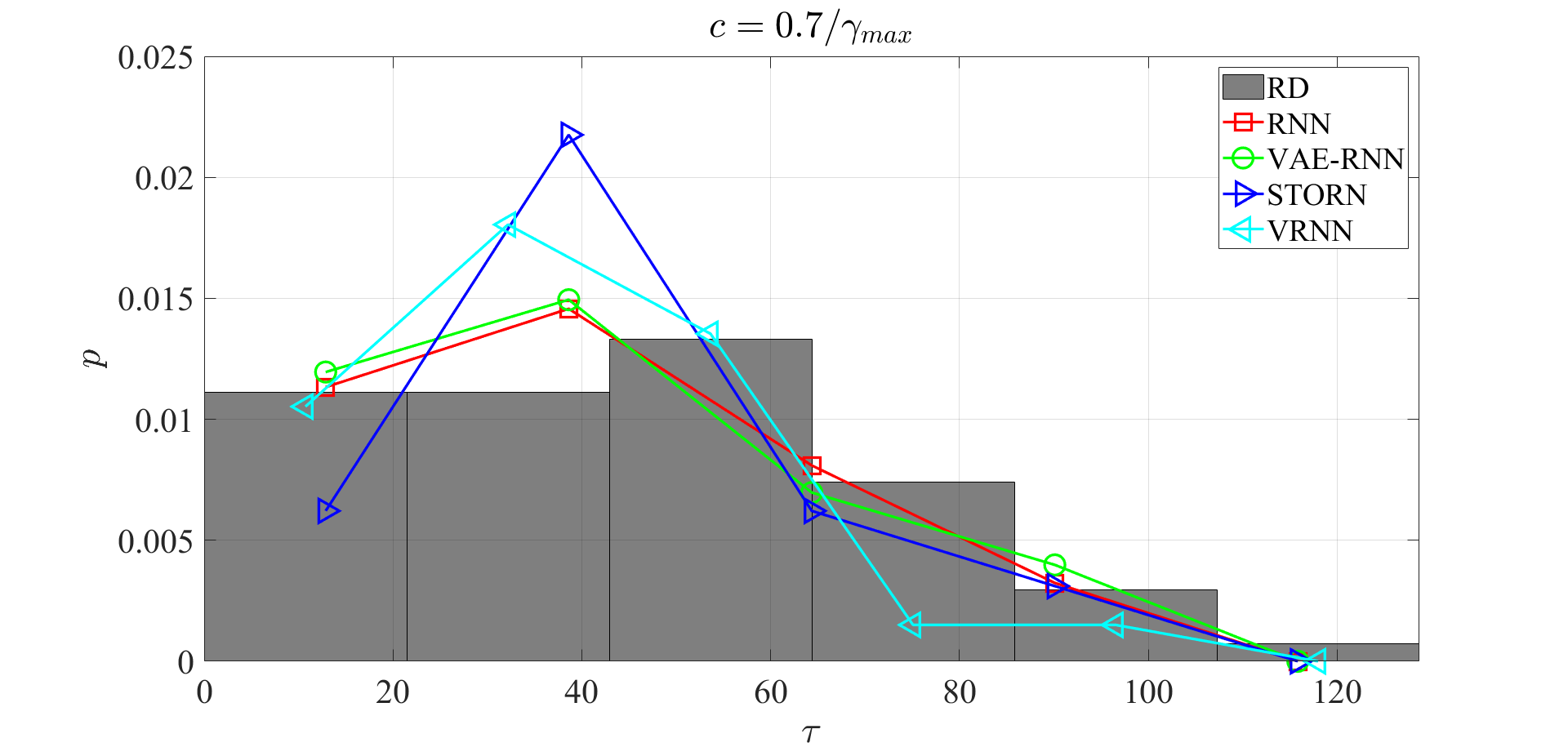} 
           \includegraphics[width=0.5\textwidth]{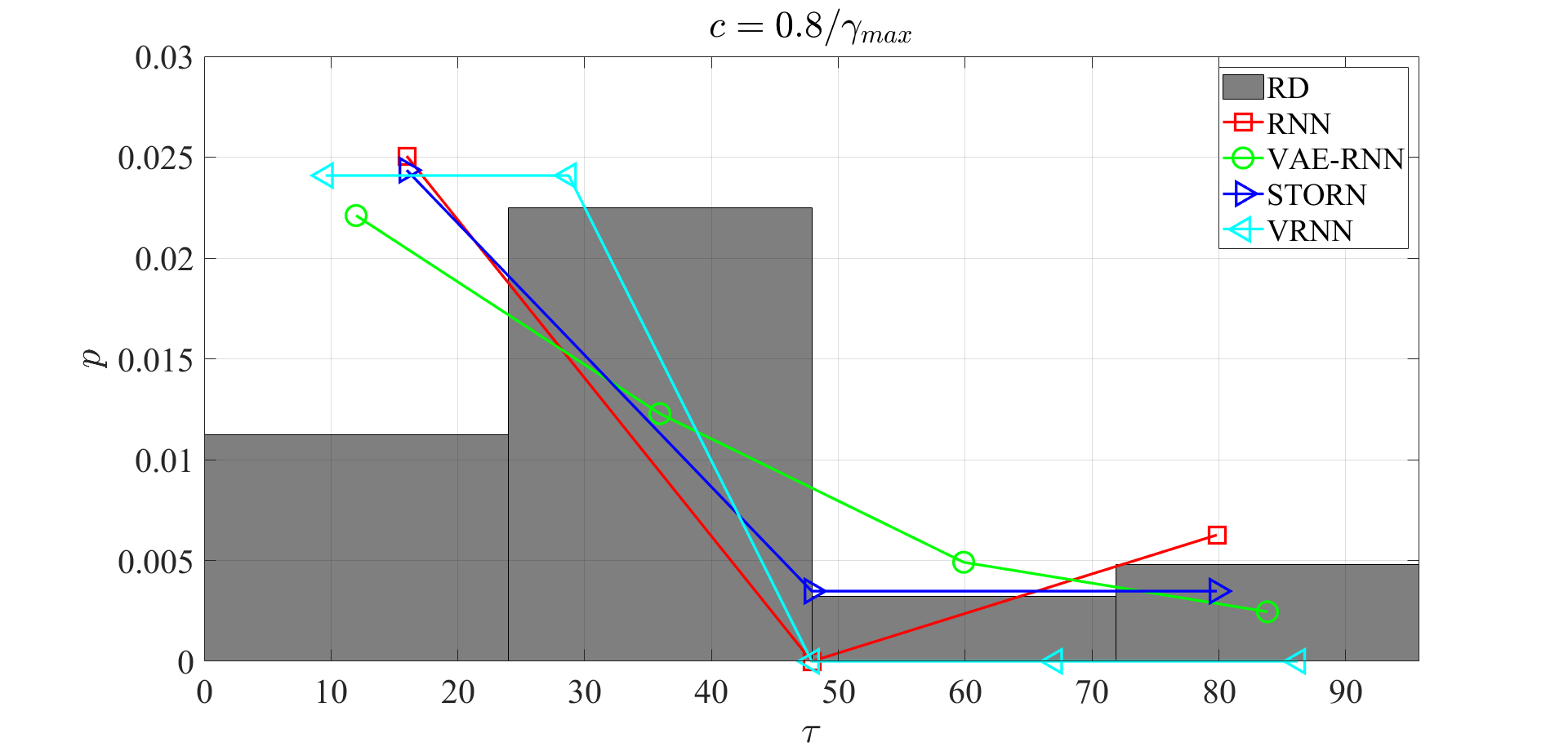} \\
              ~\ ~\  ~\ ~\ ~\ ~\ ~\ ~\ ~\ ~\ ~\ ~\ ~\ ~\ ~\ (g) ~\ ~\ ~\ ~\  ~\ ~\ ~\ ~\ ~\ ~\ ~\ ~\ ~\ ~\ ~\ ~\ ~\  ~\ ~\ ~\ ~\  ~\ ~\ ~\  ~\ ~\ ~\ (h) 
    \end{tabular}
   
    \caption{Frequency, expected duration, and variance of high amplitude excursions of $\gamma_2$ as a function of threshold $c$ (a). Probability density function of $\tau$ for fixed values of $c$ (b-h).}
    \label{fig:gamma_stats}
\end{figure}

In the context of climate, persistent extreme weather events such as long periods of high temperature (heat waves) or low precipitation (droughts) can have outsized effects on the population \cite{perkins_kirkpatrick_2020}. In order to implement effective mitigation strategies it is crucial to accurately quantify the expected duration of such events, especially as these can occur over a wide range of time scales from days to months (heatwaves) or years (droughts).  These concerns are heightened by the expectation that climate change will lead to an increase in both the frequency and severity of such events \cite{barriopedro_hot_2011,geirinhas_recent_2021, meehl2004}. For these reasons, it is critical that the ML corrected flow fields accurately reflect the frequency and duration of such extended high amplitude events. While the QG model under investigation here lacks temperature or precipitation, we aim to quantify this ability through the observable 
\begin{equation}\label{eq:gamma}
    \gamma_j(y,t) = \textrm{MA}_{100}\left( |\bar{\psi}_j(y,t)|^2 \right),
\end{equation}
which we generically refer to as ``energy''. Here $\textrm{MA}_T$ represents a moving average with a window of length $T$. We use the filtered energy to eliminate high frequency fluctuations and focus instead on large deviation events which occur over long time scales.  
To quantify the statistics of high amplitude excursions of $\gamma_j(y,t)$, we count and measure the duration of periods over which the energy exceeds a given threshold $c$. We denote the duration of each such period as $\tau$. Figure \ref{fig:gamma_stats}a shows the total number $N_c$ of high amplitude periods as well as their mean duration, $\bar{\tau}$, and standard deviation $\sigma{\tau}$ as a function of threshold $c$. We consider values of $c$ ranging from $20\%$ to $90\%$ of the maximum value of $\gamma$ observed in the reference dataset: $\gamma_{max}$. Note that the uncorrected solution (CR) fails to accurately capture any of these statistics. On the other hand, all four ML predictions accurately reflect the dependence of the high amplitude excursion statistics on the threshold $c$, while slightly under-predicting the total number and average duration. However, in all cases the variance of the high amplitude excursions is well predicted. Note that the ML predictions even capture the non-monotonic behaviour of the total number of excursions $N_c$ for $0.2<c<0.4$. 
This slightly counter-intuitive behaviour indicates that the energy often fluctuates about elevated levels before decaying back down to a lower baseline. We also show in Figures \ref{fig:gamma_stats}b-h the probability density functions of the duration $\tau$ for a range of $c \in [0.2,0.8]\gamma_{max}$ -- for higher values of $c$ there are insufficient excursions for meaningfully converged statistics. We omit the pdfs of the uncorrected (CR) solution for $c>0.2\gamma_{max}$ as these fail to capture the true distributions entirely. Consistent with Figure \ref{fig:gamma_stats}a, we see that in general the pdfs of the ML predictions peak at slightly lower $\tau$ for values of $c/\gamma_{max} > 0.3$. However, the probabilistic architectures are in some case able to ameliorate this underprediction -- as seen in \ref{fig:gamma_stats}e,f. In these cases, the inclusion of the probabilistic latent space pushes the ML prediction slightly towards higher values of $\tau$ -- and thus closer to the truth. 

\section{Discussion}\label{sec:Disc}
In this work we developed a non-intrusive data-driven framework for probabilistically debiasing under-resolved long-time climate simulations. This framework, based on training a NN correction operator on nudged simulations of an under-resolved dynamical system, enables learning the intrinsic system dynamics from very short training data sets. \textcolor{black}{The probabilistic extension we propose in this work allows us to significantly improve the extrapolation capabilities of the previous state-of-the-art and enable the quantification of the uncertainty therein.} As a test case we considered a two-layer quasi-geostrophic flow in a periodic domain with imposed bottom topography. \textcolor{black}{The topography was included to introduce anisotropy for the purposes of studying the ability of our approach to capture varying regional statistics -- a feature not included in the QG example described in previous work. The ML correction operators were trained on trajectories spanning 1,000 time units and tested on 34,000 time units -- the statistics of which differ significantly from those of the much shorter training data.} \textcolor{black}{We demonstrated the superior performance of our probabilistic framework through it's increased ability to accurately predict both global and regional statistics as well as multiple metrics quantifying the spatial and temporal distribution of rare events. The improvements over the deterministic model described in previous work were especially pronounced when analyzing the spatial variation of extreme (high amplitude) events -- a crucial feature in assessing the impact of extreme weather.}

\textcolor{black}{One of the key innovations of this work is the variational generalization of the LSTM based network architectures used in previous studies. } We investigated three recently proposed architectures (VAE-RNN, STORN, VRNN) \cite{fraccaro_deep_2018,bayer_learning_2015,chung_recurrent_2016}, which primarily differ in the way the probabilistic latent space interacts with the recurrent layer of the network. These dependencies can be categorized as being either \textit{upstream} or \textit{downstream} of the recurrence relation in the computational graph. While we found that all three architectures provide a benefit over the deterministic baseline, the VAE-RNN, which has a strict downstream dependence, achieved the lowest overall error as measured by the KL-divergence. However, the downstream dependency hinders the prediction of outlier events and leads to an overestimation of the high frequency spectral content. These issues are ameliorated through the introduction of an upstream latent space dependency which further regularizes the latent space by allowing for communication between time steps of the latent space encoding. Accordingly, the STORN (upstream only) and VRNN (upstream and downstream) architectures demonstrate the greatest ability to accurately capture the far tails of the true distribution as well as the energy content across the full spectral range. Additionally, we found the STORN and VRNN architectures were significantly more robust to over-fitting, with the VAE-RNN on the other hand showing significant deterioration in predicative capabilities when trained for longer than optimal. However, as the VAE-RNN is simply a VAE appended independently at each time step, these shortcomings should be weighed against its simplicity and ease of implementation. \textcolor{black}{The optimal architecture design will likely depend on the specific application and we hope the analysis discussed in this work can serve as a guide to researchers employing our framework.}

While our approach has demonstrated significant skill in correcting the long time statistics of the QG climate model over a range of scales, several limitations remain. Specific to our results: the accurate reconstruction of two point statistics remains a challenge. Our results show that the primary means by which our approach corrects the under-resolved trajectory is by correcting the spatiotemporal dynamics of different Fourier modes independently -- while the phase shifts between these remain relatively unchanged. One potential avenue to address this issue is through the use of Fourier Neural operators \cite{li_fourier_2021,li_fourier_2022,pathak_fourcastnet_2022} which operate directly in Fourier space, and may therefore be more effective at correcting the small discrepancies in phase shifts between individual modes. Additionally, the ML corrected fields slightly, but systematically, underestimate the number and duration of high amplitude excursions. \textcolor{black}{Additionally, in order to facilitate long time horizon simulations we have focused in this work on a simplified flow, and quantifying the improvements of our probabilistic framework when applied to a full-scale climate model remains the topic of ongoing research.}

The proposed framework also has several more fundamental limitations which must be mentioned. First and foremost, an intrinsic limitation of post-processing approaches is their inability to correct processes that are missing from the coarse-resolution model entirely. Improvements in the representation of such processes, such as cloud formation and convective precipitation, requires intrusive corrections to the coarse-resolution model via either improved subgrid-scale closures \cite{schneider_climate_2017, Cohen2020, lopezgomez_2020}, or localized high-resolution simulation \cite{Randall_2003, Kooperman_2016}. Second, the current framework implicitly assumes that the system is statistically stationary. While similar frameworks have been applied in non-stationary systems \cite{zhang_machine_2024}, a correction operator trained under this assumption may fail when applied to trajectories which include strong transitory periods. Finally, the fact that the ML correction operators discussed here are intended to produce long time statistics, but are trained on very short data, implies that there is no obvious metric which can be monitored during training to prevent over-fitting (see \ref{app:model_selection}). However, we have found that the upstream latent space dependencies, as in the STORN and VRNN architectures, serve to regularize the network and drastically increase the robustness of the NN's to over-training. Additionally, ensemble-based predictions help to ameliorate these concerns even further. However, we acknowledge that the efficacy of these strategies may vary from application to application, and in some cases more rigorous regularization strategies may be needed.

\textcolor{black}{In conclusion, we have demonstrated that ensembles of VAE-based RNNs are effective at increasing the extrapolation and rare event quantification capabilities of the non-intrusive debiasing framework introduced by \citeA{barthel_sorensen_non-intrusive_2024}.} We investigated several recently developed architectures, which differ primarily in how the probabilistic latent space interacts with the recurrent layer of the neural network. We have classified these interactions as upstream or downstream, and demonstrated that while both are effective, networks with downstream interactions -- especially in the absence of additional upstream interactions --  are susceptible to over-fitting, and noise corruption. While our work has focused on the application to climate modeling, the general training strategy outlined in this work is applicable to any scenario in which long time statistical analysis requires computationally intractable high resolution numerical simulations. 
\section{Acknowledgments}
This work has been supported through the Google-MIT program ``Hybrid Physics and Data-Driven Methods for Statistics of Extreme Weather Events from Climate Simulations”.

\section*{Open Research Section}
The software and data needed to generate the results described in this work can be found at
\url{https://zenodo.org/doi/10.5281/zenodo.13833073} \cite{barthel_data_2024} and \url{https://github.com/ben-barthel/learning_dynamics}.
\appendix
\section{}
\subsection{Validation and Model Selection}\label{app:model_selection}
Many machine learning applications operate in what might be referred to as a ``data-rich'' environment. Even if the total available data is small in an absolute sense, it is common to use a large fraction $75-90\%$ of this available data for training, with only the small remainder used to generate the presented results. The ML models considered here operate in a much more ``data-poor'' environment, with only $3\%$ of the total data seen in training. 
One of the challenges in this regard is the lack of obvious metric for online validation. In a ``data-rich'' environment, a small fraction of the training set may be set aside for validation. Then, as training progresses, the validation error, i.e. the training loss evaluated on the validation set, is monitored and training is stopped when the validation error no longer decreases with each passing epoch. However, if the goal is statistical accuracy over time horizons much longer than the training data, monitoring the training loss (generally the L2 error) over a small fraction of the already limited training set does not provide meaningful insight into the eventual performance when applied to long time series data. 

\begin{figure}
        \centering
        \begin{tabular}{ll}
        \includegraphics[width=0.95\textwidth]{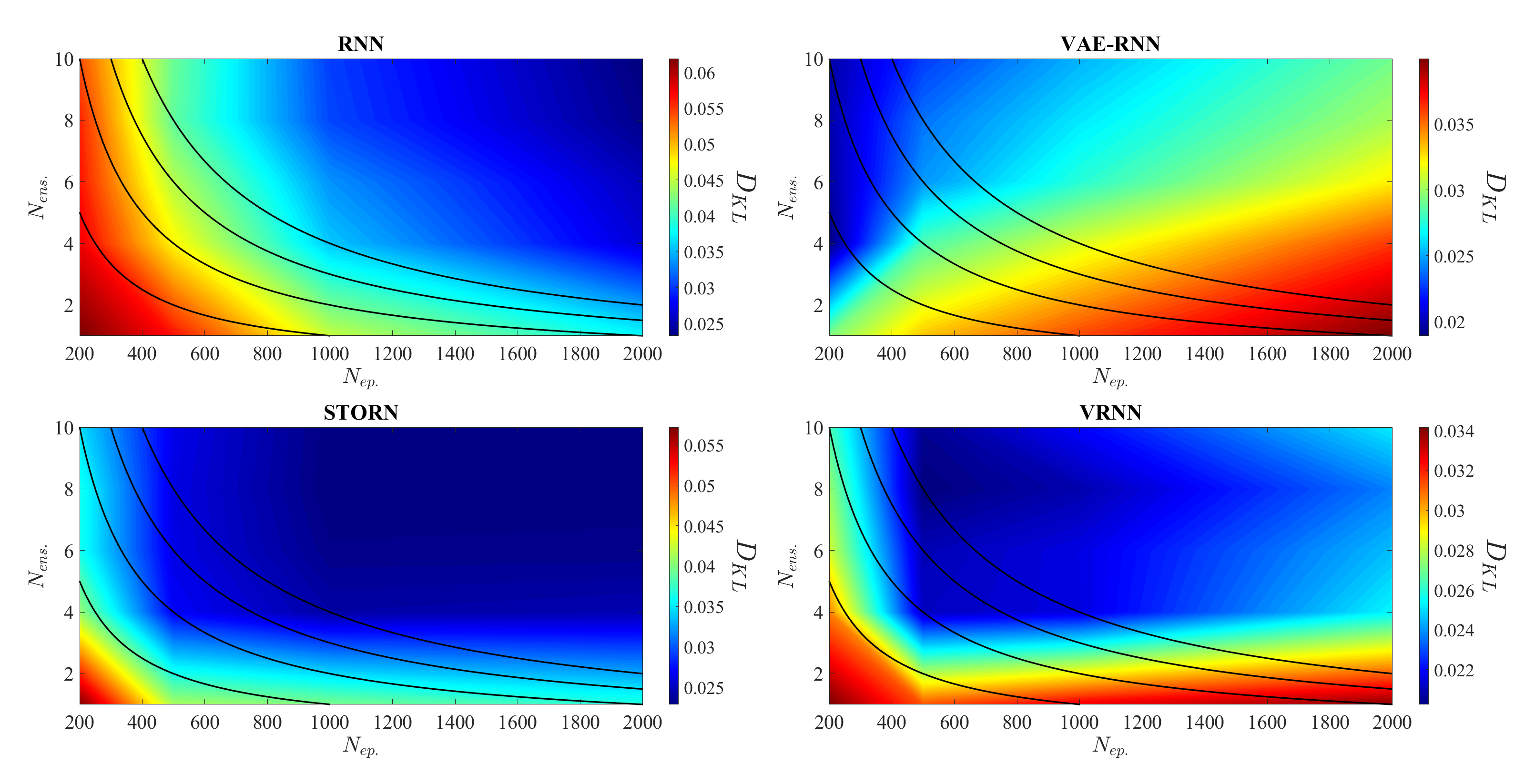} \\
         ~\  ~\ ~\  ~\ ~\  ~\ ~\  ~\  ~\  ~\ ~\  ~\ ~\  ~\ ~\  ~\ ~\  ~\  ~\   ~\   ~\   ~\  ~\  ~\  ~\   ~\   ~\   ~\  ~\  ~\  ~\   ~\   ~\  (a)   \\
       \includegraphics[width=0.95\textwidth]{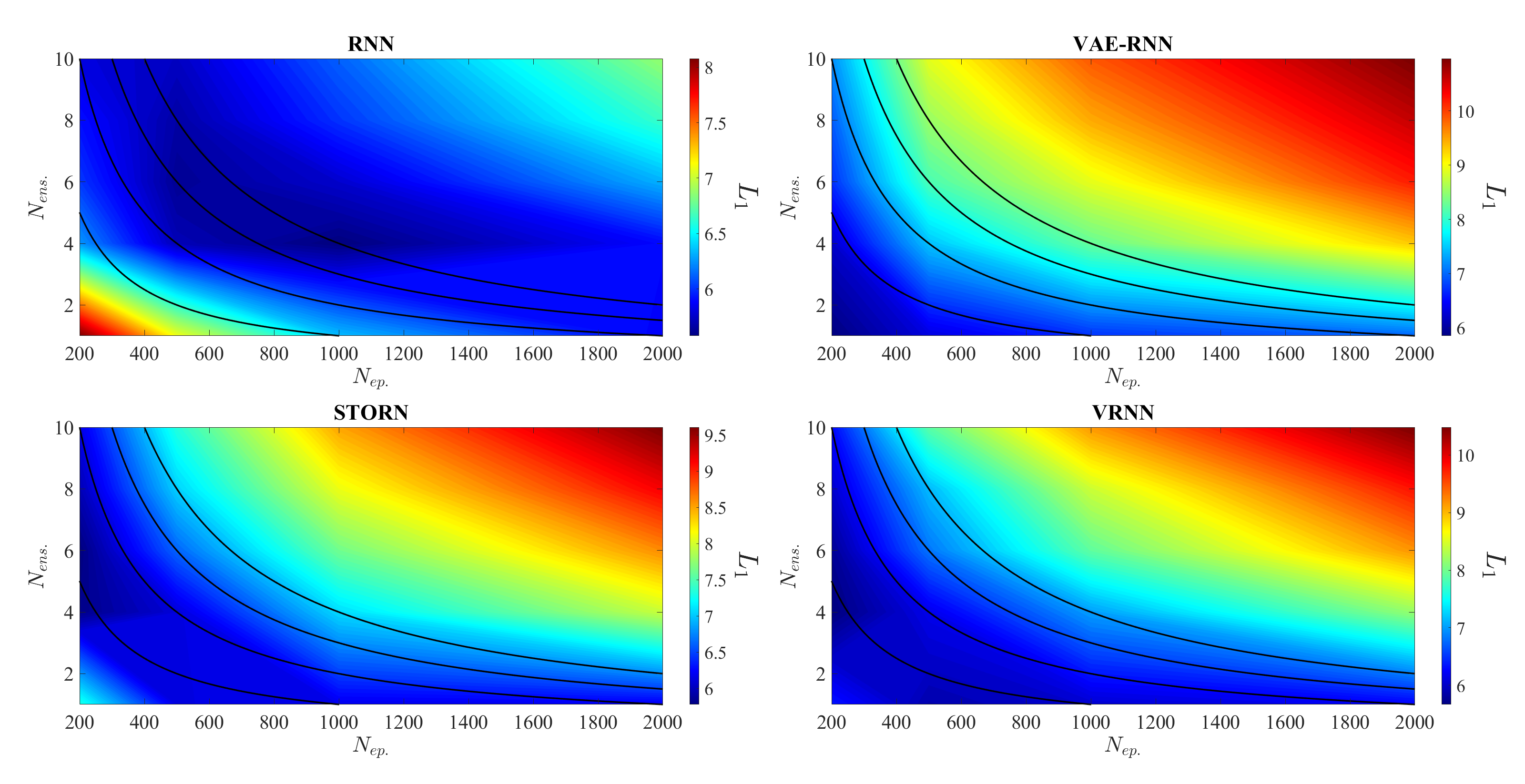}
        \\
          ~\  ~\ ~\  ~\ ~\  ~\ ~\  ~\  ~\  ~\ ~\  ~\ ~\  ~\ ~\  ~\ ~\  ~\  ~\   ~\   ~\   ~\  ~\  ~\  ~\   ~\   ~\   ~\  ~\  ~\  ~\   ~\   ~\  (b)     \\
        
        \end{tabular}
        \caption{Average error in prediction as a function of ensemble size and number of epochs trained. KL divergence (a), L1 norm of log pdf error (b), integrated variance (c), and integrated log variance (d). Lines indicate 1000, 2000, 3000, and 4000 total epochs (left to right).}
        \label{fig:error_eps_Nens_psi}
\end{figure}

To this end, we conducted a parametric study of the impact of both training time per ensemble member and ensemble size. The results thereof are summarized in figure \ref{fig:error_eps_Nens_psi} which shows the global average (over $\psi_1$ and $\psi_2$) KL-divergence \eqref{eq:KLD} and $L_1$ error \eqref{eq:L1}. This parametric study revealed four crucial observations. First, the probabilistic architectures generally lead to lower Kl divergence regardless of training time or ensemble size -- note the different color scales between the four subfigures in figure \ref{fig:error_eps_Nens_psi}a.  Second, for a given ensemble size the variational models require less training time to reach a desired level of accuracy. Third, above a certain minimum training time -- approximately 500 epochs -- it is more advantageous to increase the ensemble size rather than train the models for longer. 
Fourth and finally, for the probabilistic architectures the error in the prediction of the tails (quantified by the $L_1$ metric) increases if the model is trained for too many epochs. This is especially pronounced for the  VAE-RNN architecture, and is in contrast to the KL-divergence -- which quantifies the overall accuracy -- which decreases monotonically in almost all cases. The lone exception being the small ensembles of the VAE-RNN architecture.

This deterioration of rare event prediction with increased training is likely due to the probabilistic models over-fitting to the latent space prior. The magnitude of the MSE term in the training loss is proportional to the magnitude of the model output, while the KL divergence term enforcing the latent space prior remains the same order of magnitude regardless of the output. This means that the optimization will tend to ignore errors in the tails of the output distribution in favor of driving the latent space representation of these outlier events ever closer to the pure Gaussian prior. This phenomenon is especially pronounced, in the VAE-RNN with its purely downstream latent space dependency \eqref{eq:rnn_vae_rnn}. In that case -- assuming a linear activation -- we have $y_t\sim z_t$ and thus the model output will become increasingly corrupted by white noise. 
This mechanism is also present to a smaller extent in the VRNN architecture, but is largely ameliorated by the regularizing affects of the upstream latent space dependency which enables communication between time steps $z_t$.

\begin{figure}
    \centering
    \includegraphics[width=0.95\textwidth]{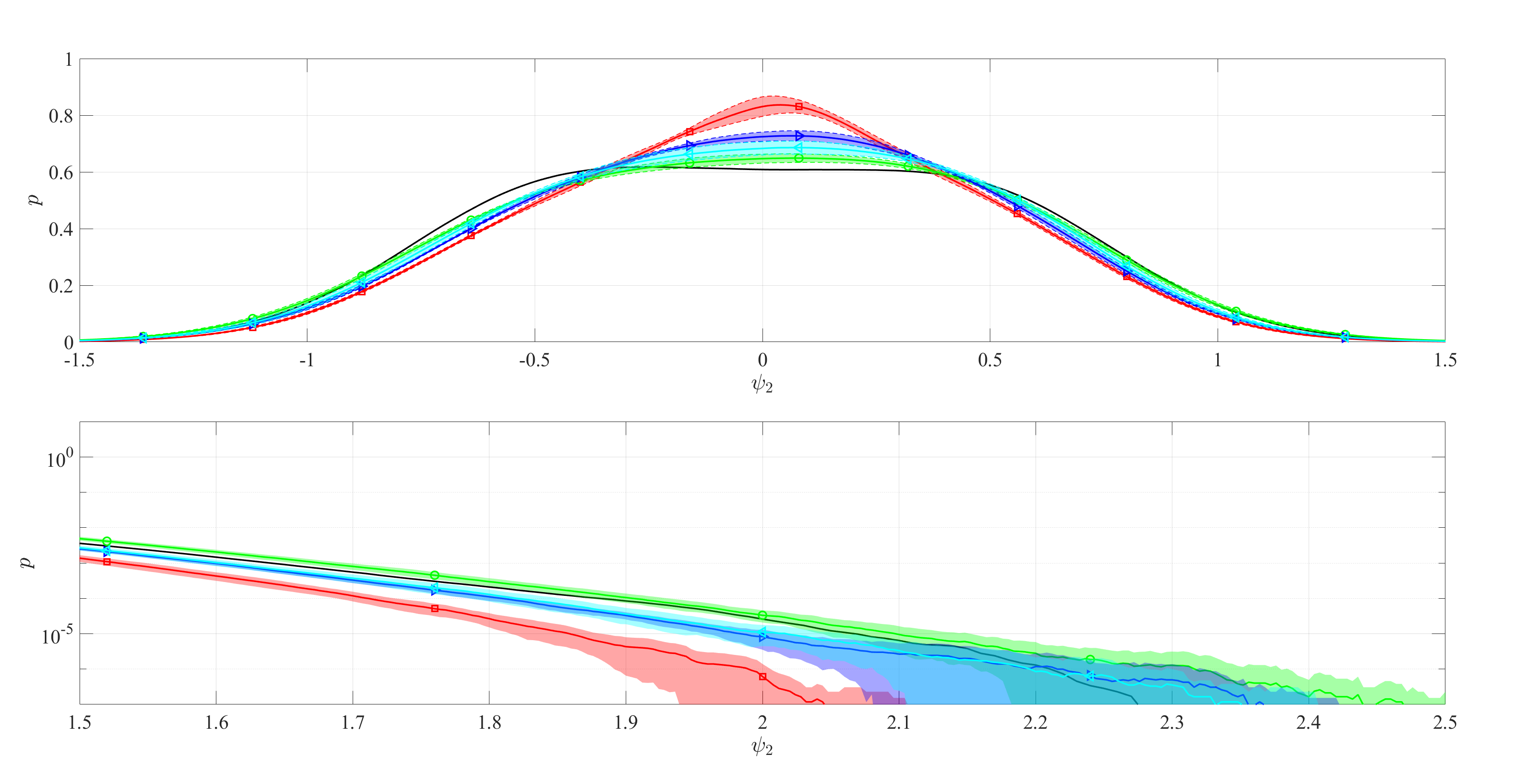}
    \caption{Global pdf of $\psi_2$ on a linear scale  (upper panel) and log-scale (lower panel). RD (solid black),  RNN (red), VAE-RNN (green), STORN (blue), VRNN (teal). Shaded area signifies ensemble mean  $\pm$ 1 standard deviation.}
    \label{fig:app_pdf}
\end{figure}

To illustrate the effects of considering an ensemble of NNs we show in figure \ref{fig:app_pdf} the ensemble mean and one standard deviation spread of the global pdf predictions for an ensemble size of 6 -- the same as the results presented in \S\ref{sec:Results}. To illustrate the variance in both the bulk and the tails of the distribution we plot these on a linear and log scale -- for the latter we zoom in on the tail of the pdf to best illustrate the ensemble variance. There are two main conclusions to be drawn from this figure. First, the variance is modest but meaningful -- most notably the tails of the pdf -- indicating that an ensemble analysis does improve the predictive capabilities of the ML correction. Second, the variance is very similar across all architecture types. 

\FloatBarrier
\subsection{Additional Results}\label{app:psi_1_results}
\begin{figure}
    \centering
    \includegraphics[width=0.95\textwidth]{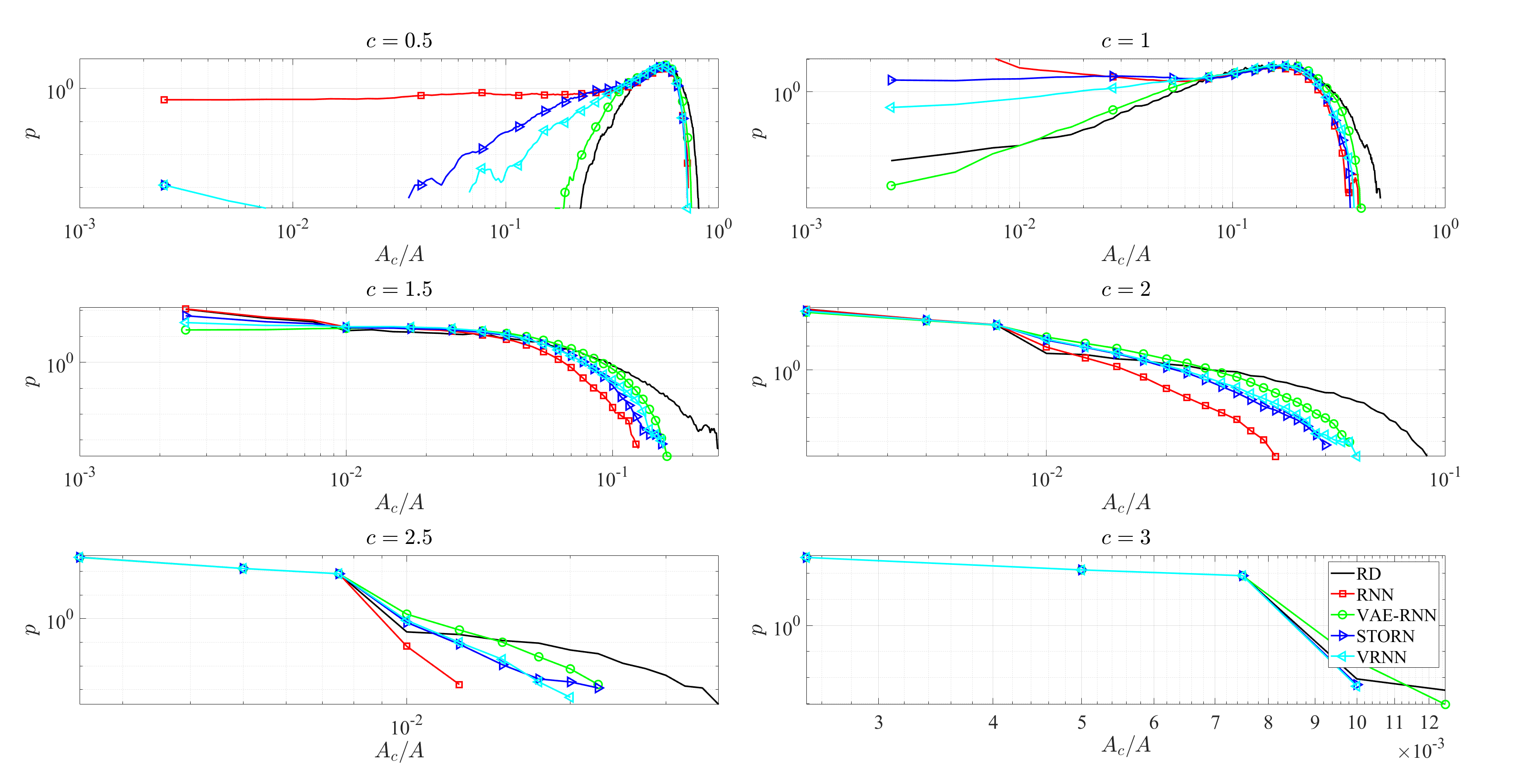}
    \caption{Pdf of fraction of domain over which $|\psi_1|$ exceeds fixed threshold $c$ for range of $c \in [0,2]$. RD (solid black), RNN (red), VAE-RNN (green), STORN (blue), VRNN (teal).}
    \label{fig:aot_psi1}
\end{figure}
\begin{figure}
    \centering
    \includegraphics[width=0.95\textwidth]{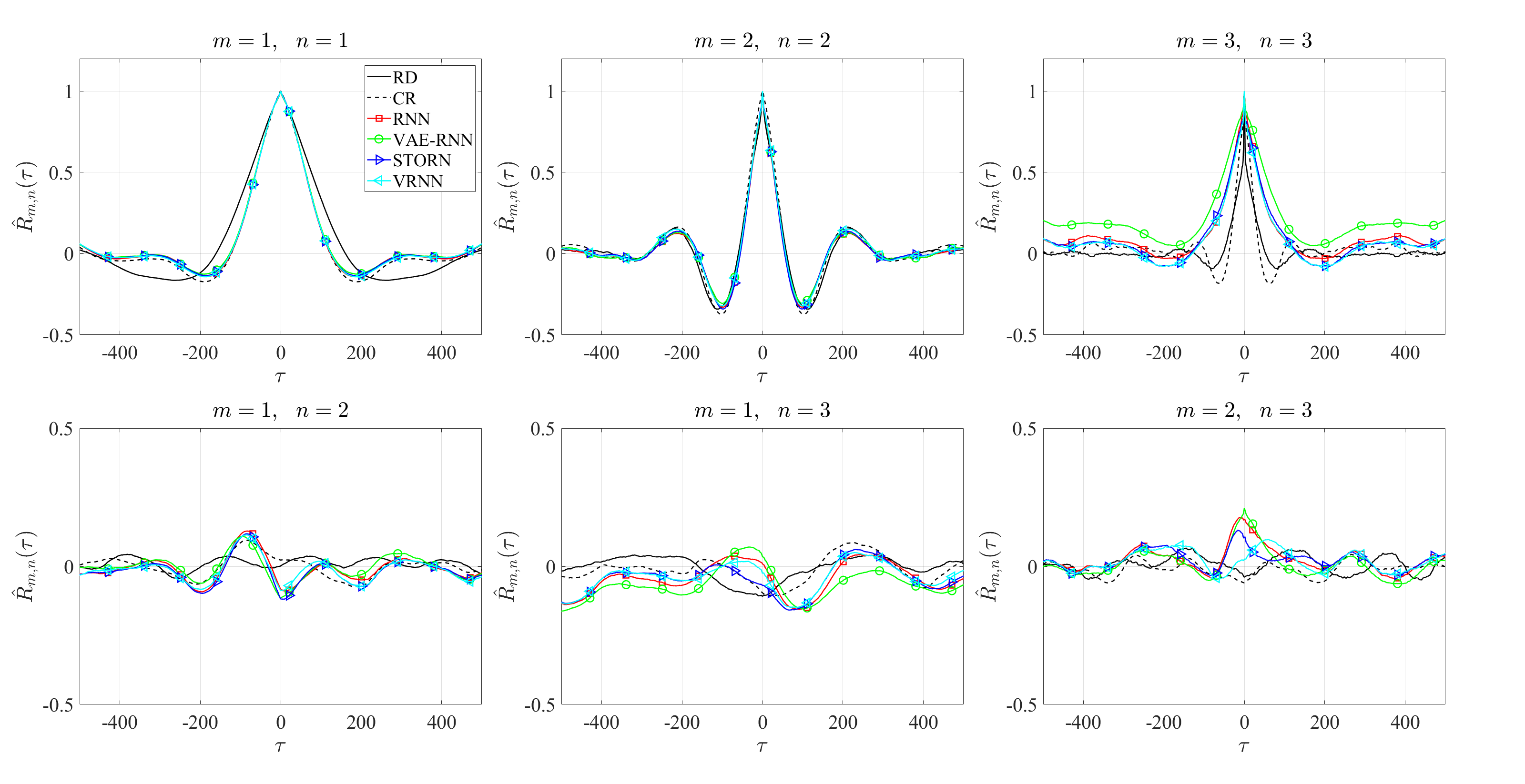}
    \caption{Normalized correlation between three largest zonally constant Fourier modes of $\psi_1$. RD (solid black), CR (dashed black), RNN (red), VAE-RNN (green), STORN (blue), VRNN (teal).}
    \label{fig:xcorr_psi1}
\end{figure}
\begin{figure}
        \centering
        \begin{tabular}{ll}
       \includegraphics[width=0.95\textwidth]{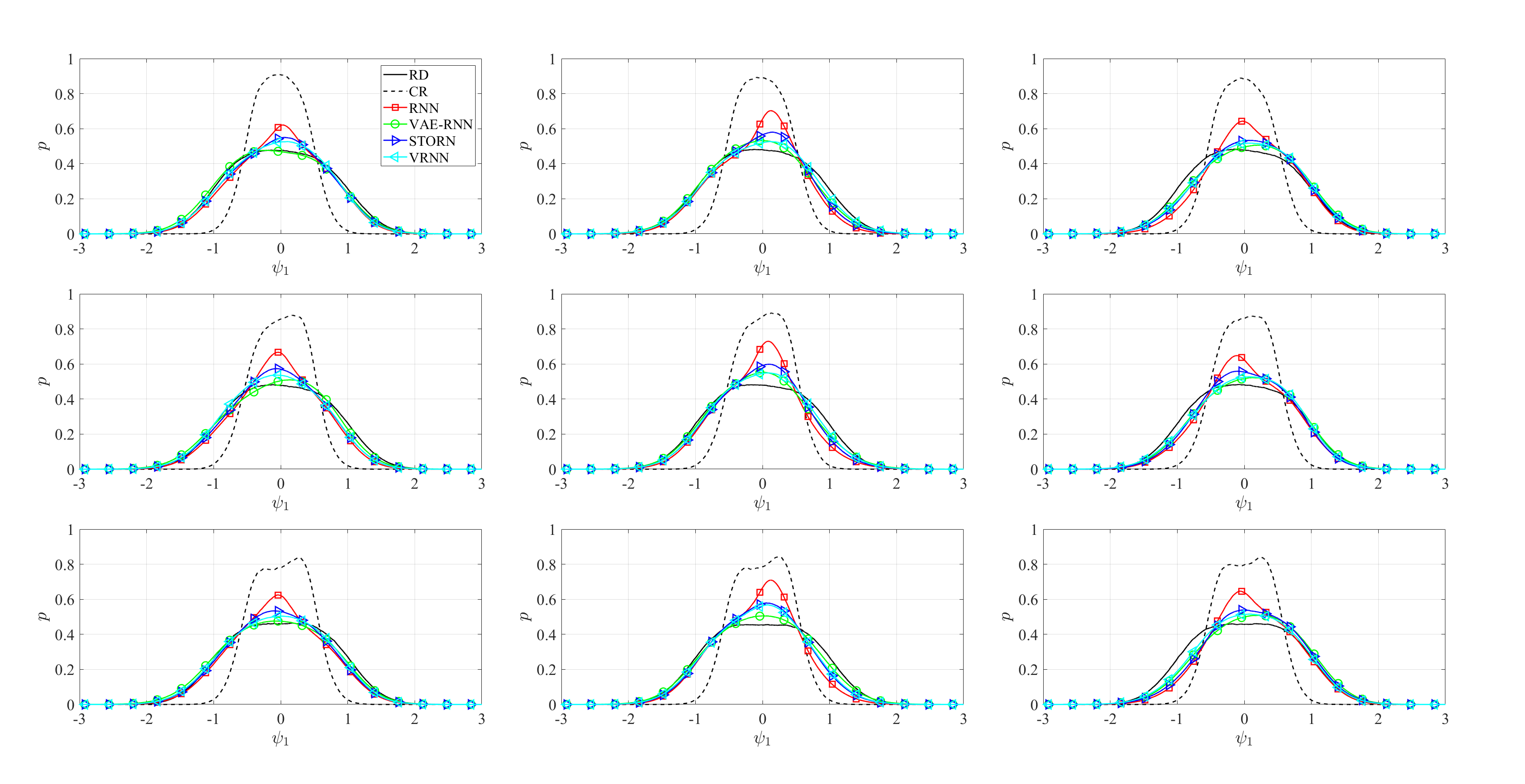}
      \\
    ~\  ~\    ~\  ~\    ~\   ~\   ~\   ~\   ~\   ~\  ~\   ~\  ~\  ~\   ~\   ~\  ~\  ~\  ~\   ~\   ~\   ~\  ~\  ~\  ~\  ~\  ~\  ~\  ~\  ~\  ~\  ~\  (a)  \\
        
        \includegraphics[width=0.95\textwidth]{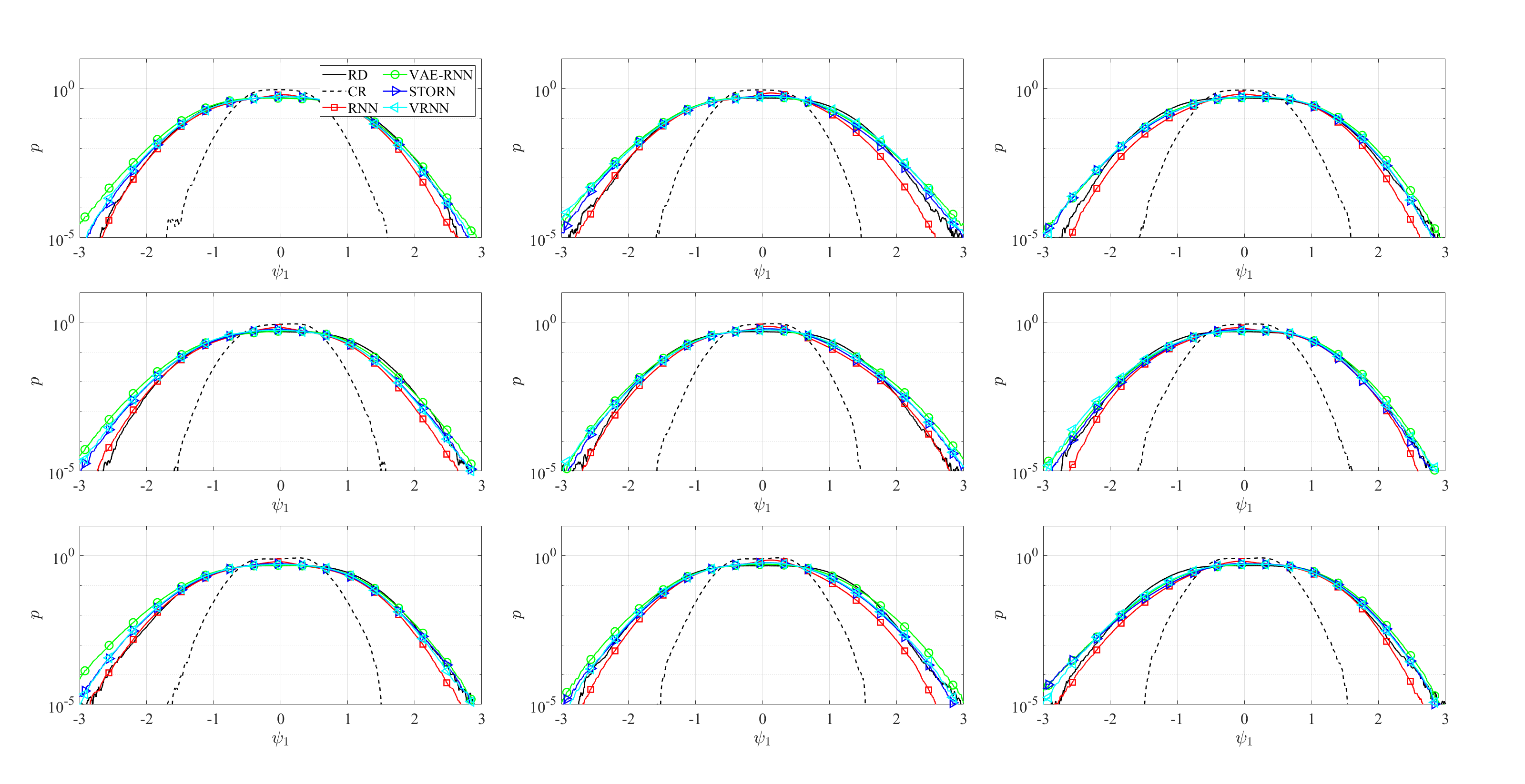}
        \\
          ~\  ~\    ~\  ~\    ~\   ~\   ~\   ~\   ~\   ~\  ~\   ~\  ~\  ~\   ~\   ~\  ~\  ~\  ~\   ~\   ~\   ~\  ~\  ~\  ~\  ~\  ~\  ~\  ~\  ~\  ~\  ~\  (b)  \\
        
        \end{tabular}
        \caption{Regional pdf (a) and log-pdf (b) of $\psi_1$.  RD (solid black), CR (dashed black), RNN (red), VAE-RNN (green), STORN (blue), VRNN (teal). }
        \label{fig:regional_pdf_psi1}
\end{figure}

\begin{figure}
    \centering
    \includegraphics[width=0.95\textwidth]{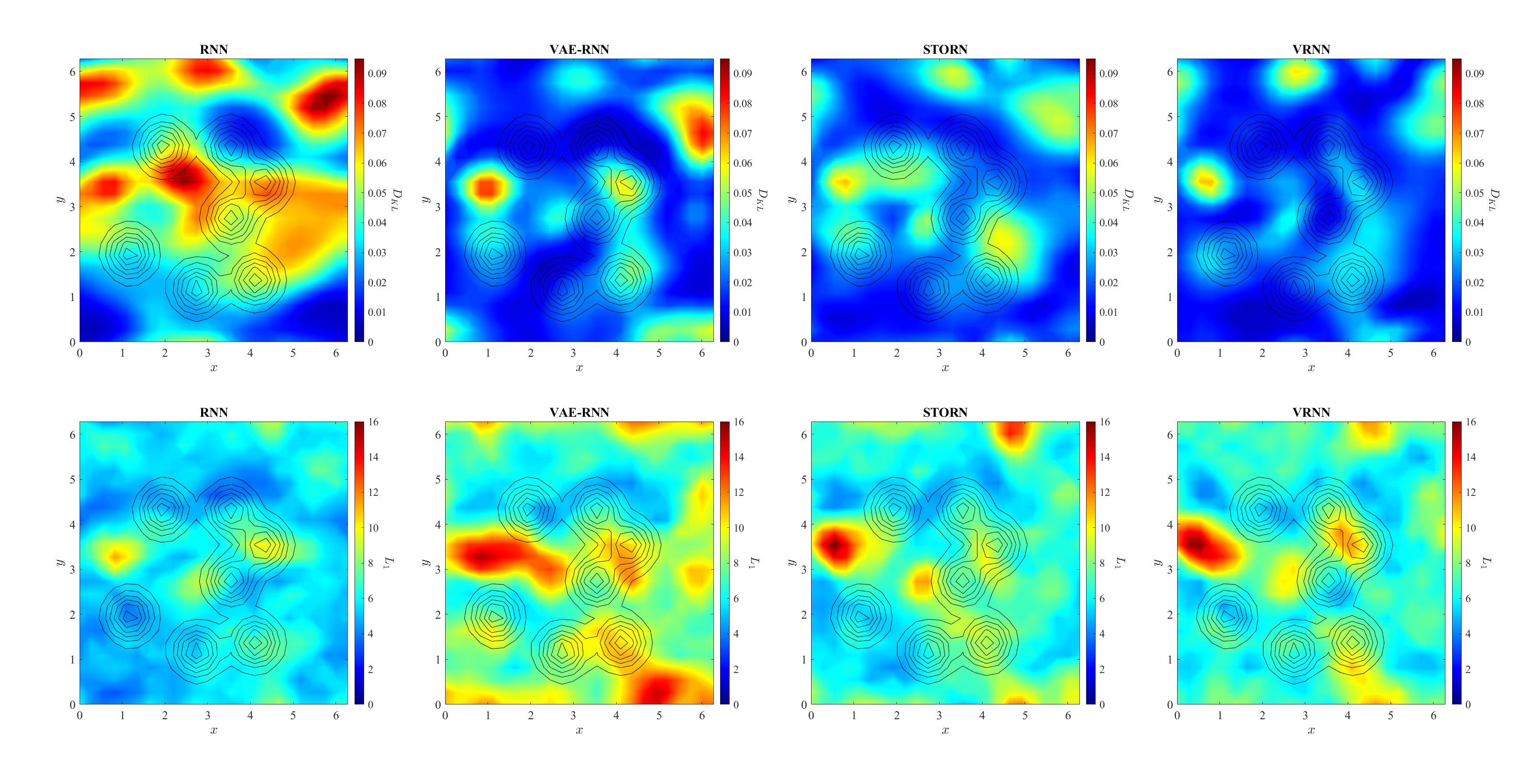}
    \caption{Spatial distribution of KL divergence (upper panel) and $L_1$ metric (lower panel) for $\psi_1$. From left to right: RNN, VAE-RNN, STORN, VRNN. The topography profile is show in black.}
    \label{fig:regional_pdf_error_psi1}
\end{figure}

\begin{figure}
    \centering
    \begin{tabular}{ll}
    \includegraphics[width=0.85\textwidth]{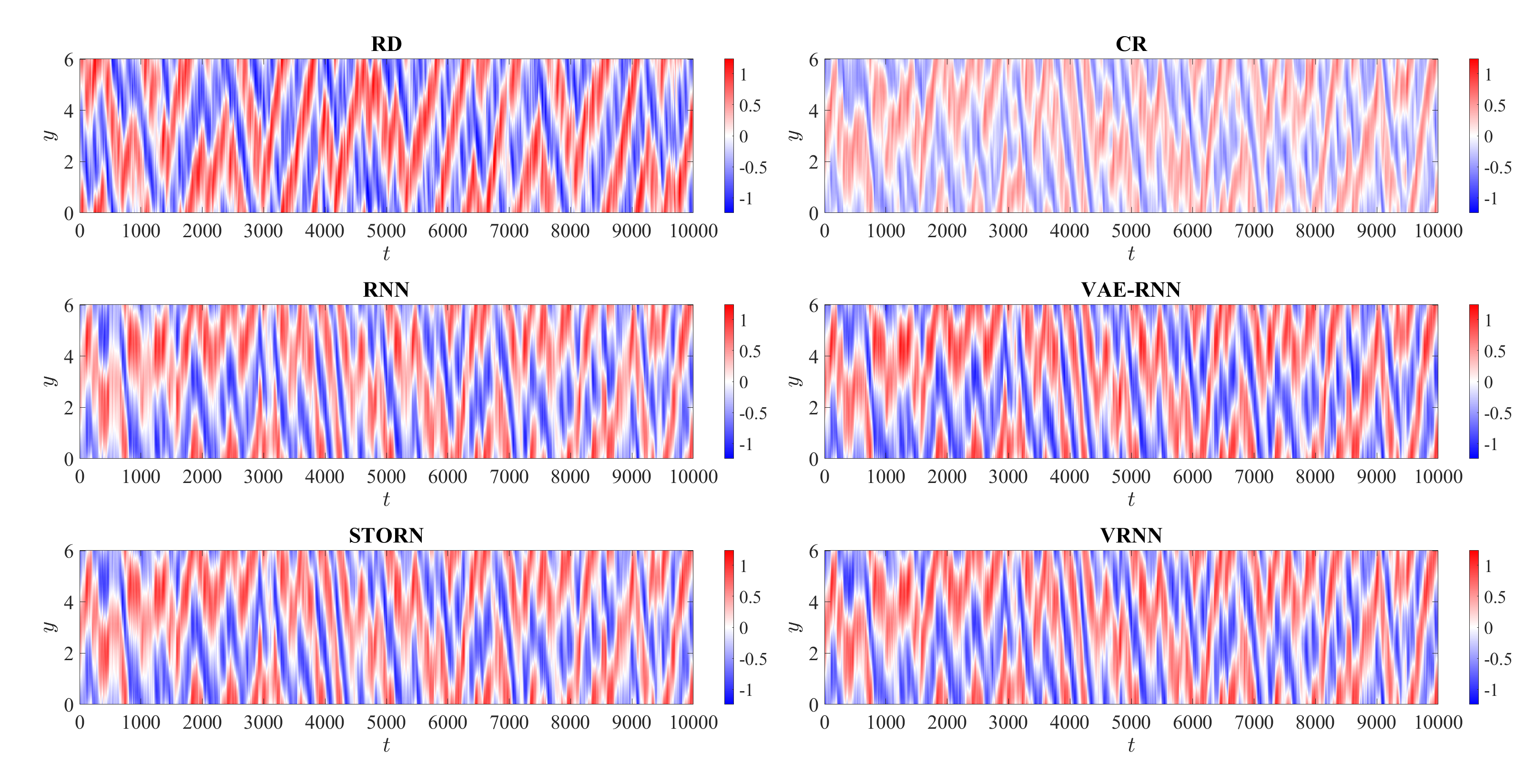} \\
    ~\ ~\ ~\ ~\ ~\  ~\ ~\ ~\ ~\ ~\ ~\ ~\ ~\ ~\ ~\ ~\  ~\ ~\ ~\ ~\ ~\ ~\ ~\ ~\ ~\ ~\ ~\ ~\ ~\ ~\ (a) \\
           \includegraphics[width=0.45\textwidth]{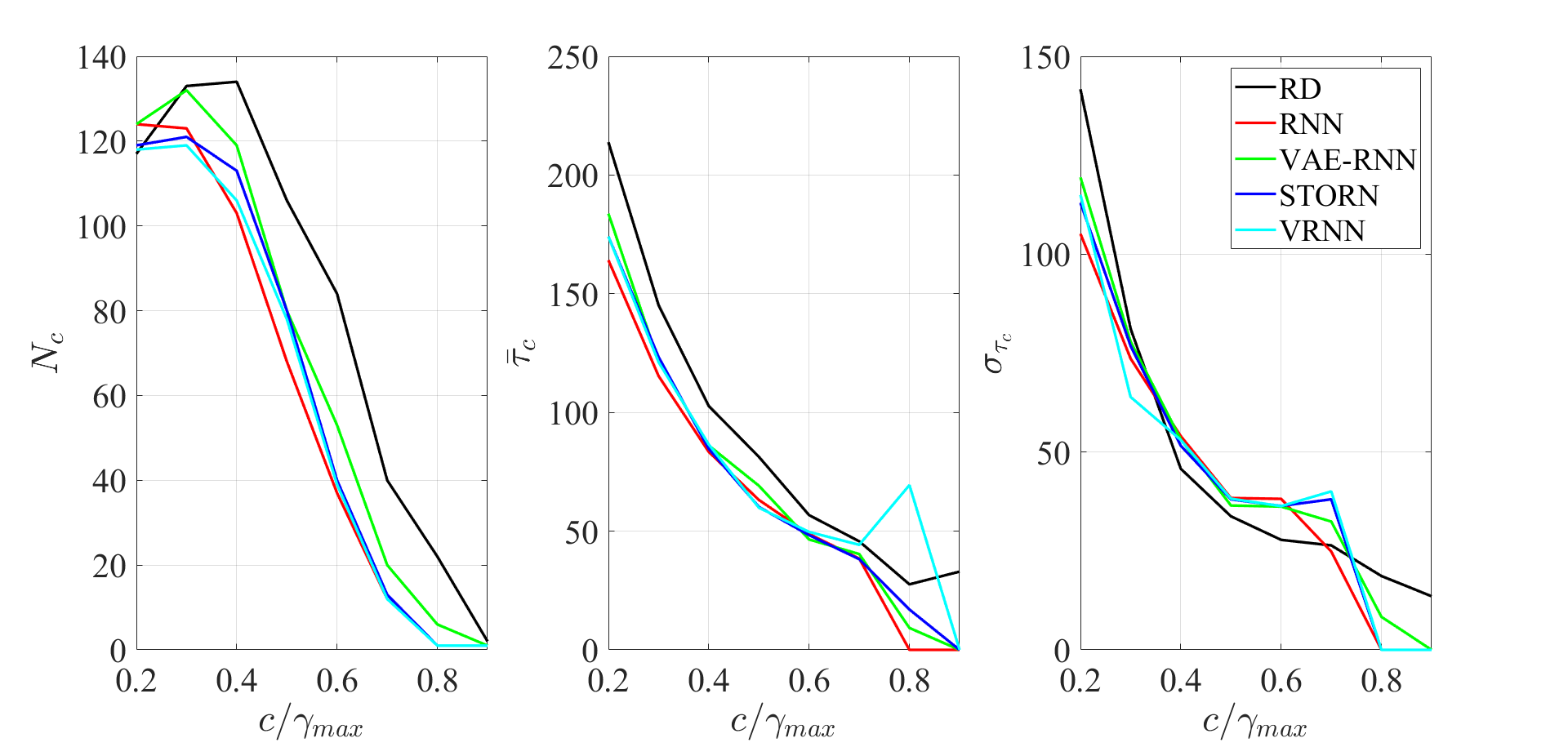}
           \includegraphics[width=0.45\textwidth]{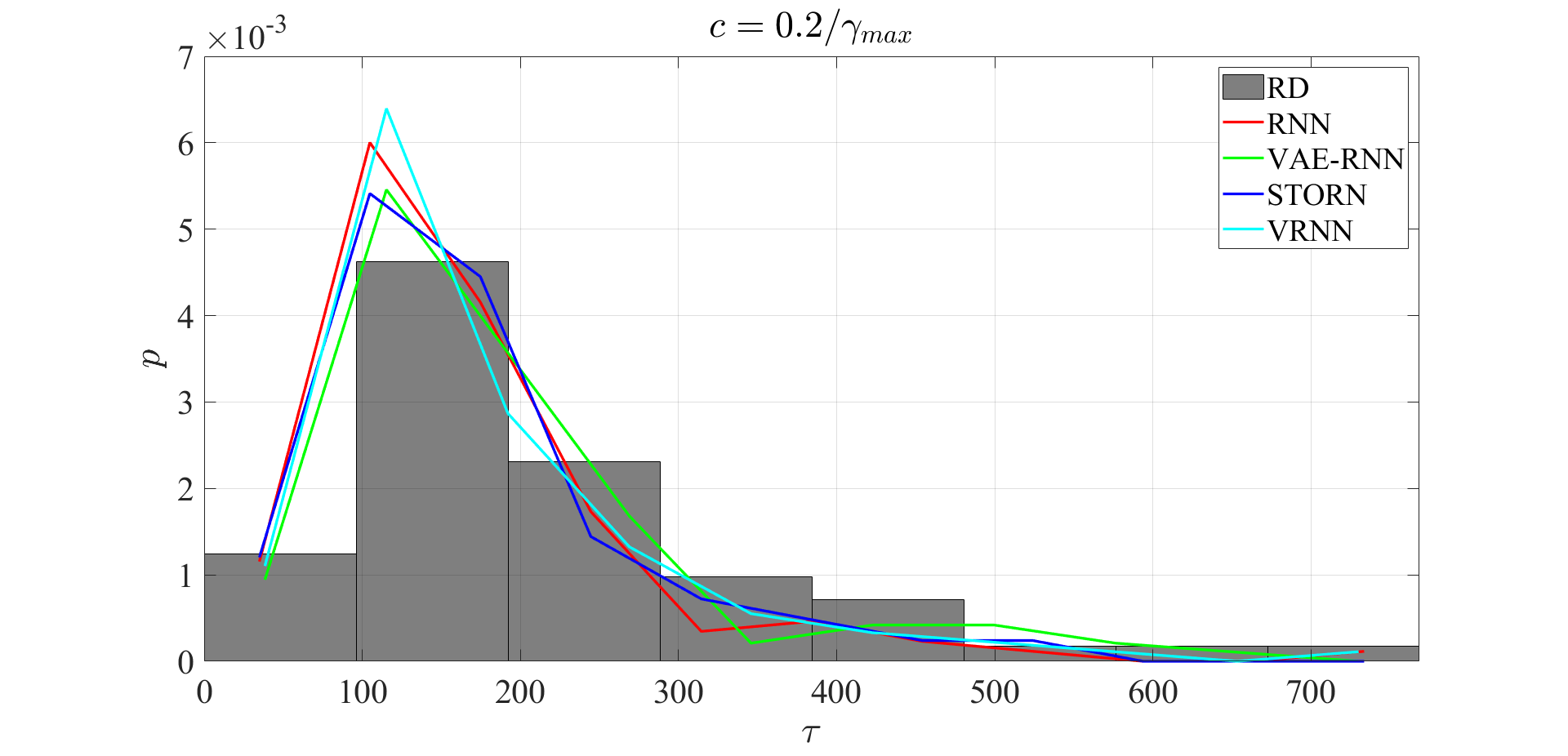} \\
             ~\ ~\  ~\ ~\ ~\ ~\ ~\ ~\ ~\ ~\ ~\ ~\ ~\ ~\ ~\ (b)  ~\  ~\ ~\ ~\ ~\ ~\ ~\ ~\ ~\ ~\ ~\ ~\ ~\ ~\  ~\ ~\ ~\ ~\  ~\ ~\ ~\ ~\ ~\ ~\ ~\ ~\  ~\ ~\ ~\ ~\ (c) \\
           \includegraphics[width=0.45\textwidth]{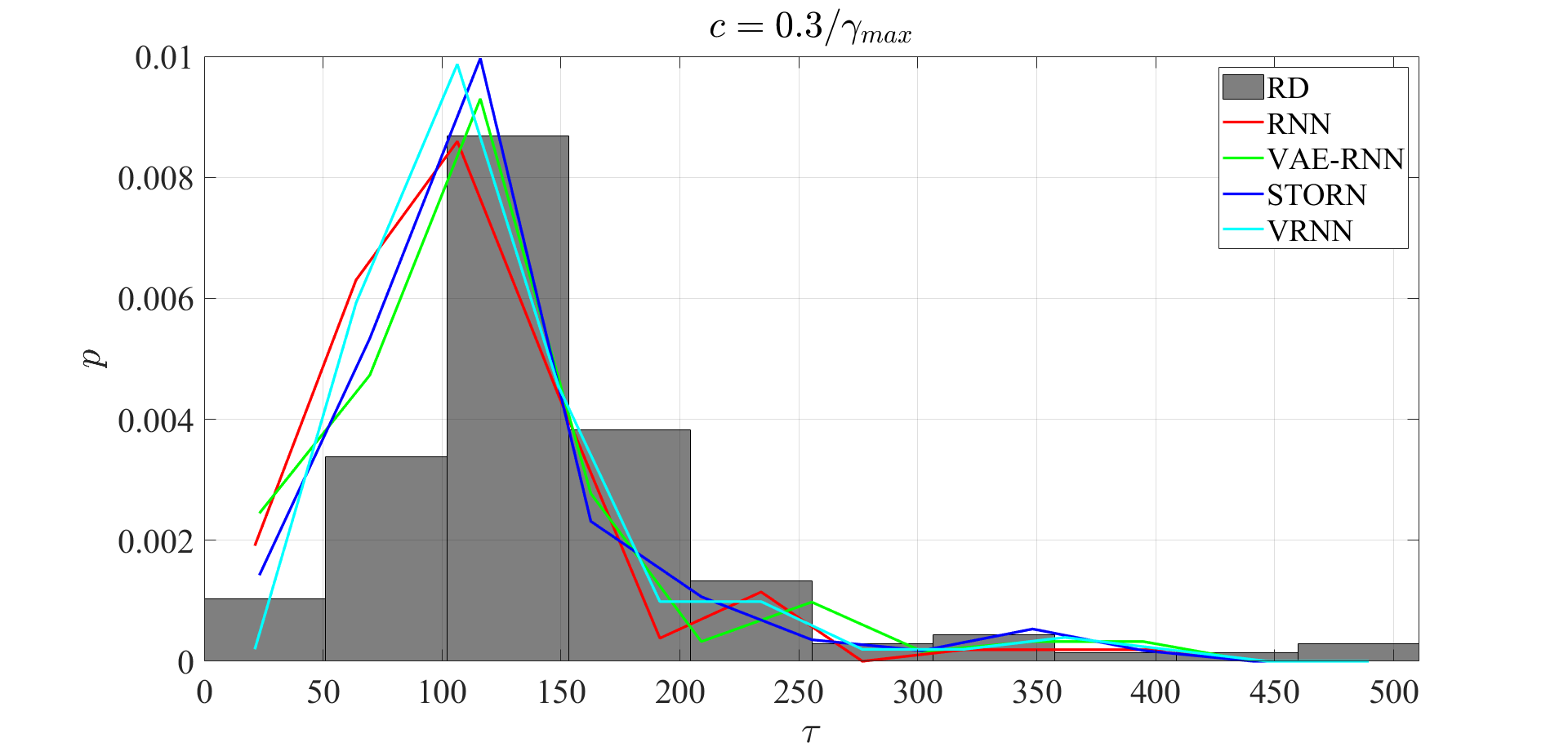} 
           \includegraphics[width=0.45\textwidth]{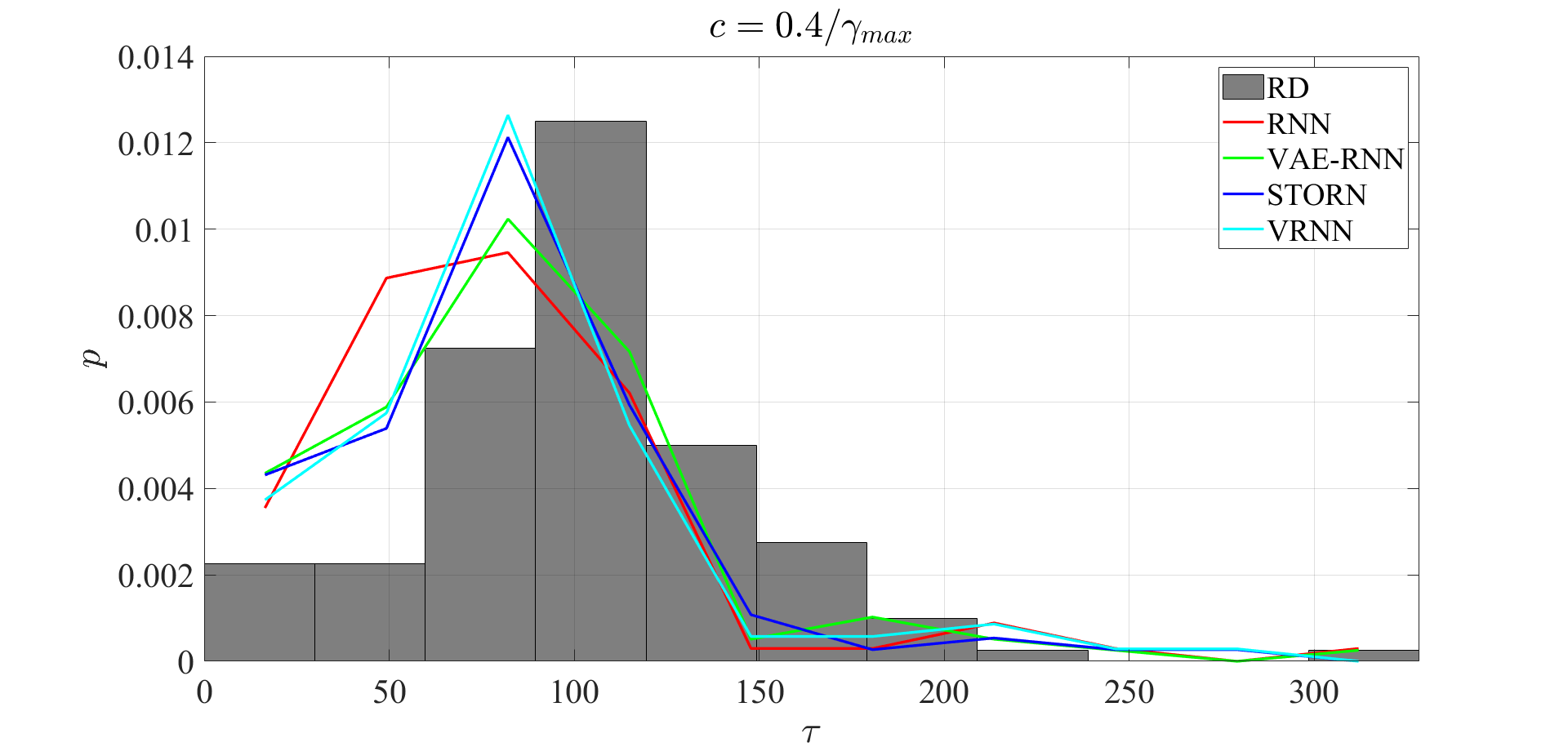} \\
             ~\ ~\  ~\ ~\ ~\ ~\ ~\ ~\ ~\ ~\ ~\ ~\ ~\ ~\ ~\ (d)  ~\  ~\ ~\ ~\ ~\ ~\ ~\ ~\ ~\ ~\ ~\ ~\ ~\ ~\  ~\ ~\ ~\ ~\  ~\ ~\ ~\ ~\ ~\ ~\ ~\ ~\  ~\ ~\ ~\ ~\ (e) \\
           \includegraphics[width=0.45\textwidth]{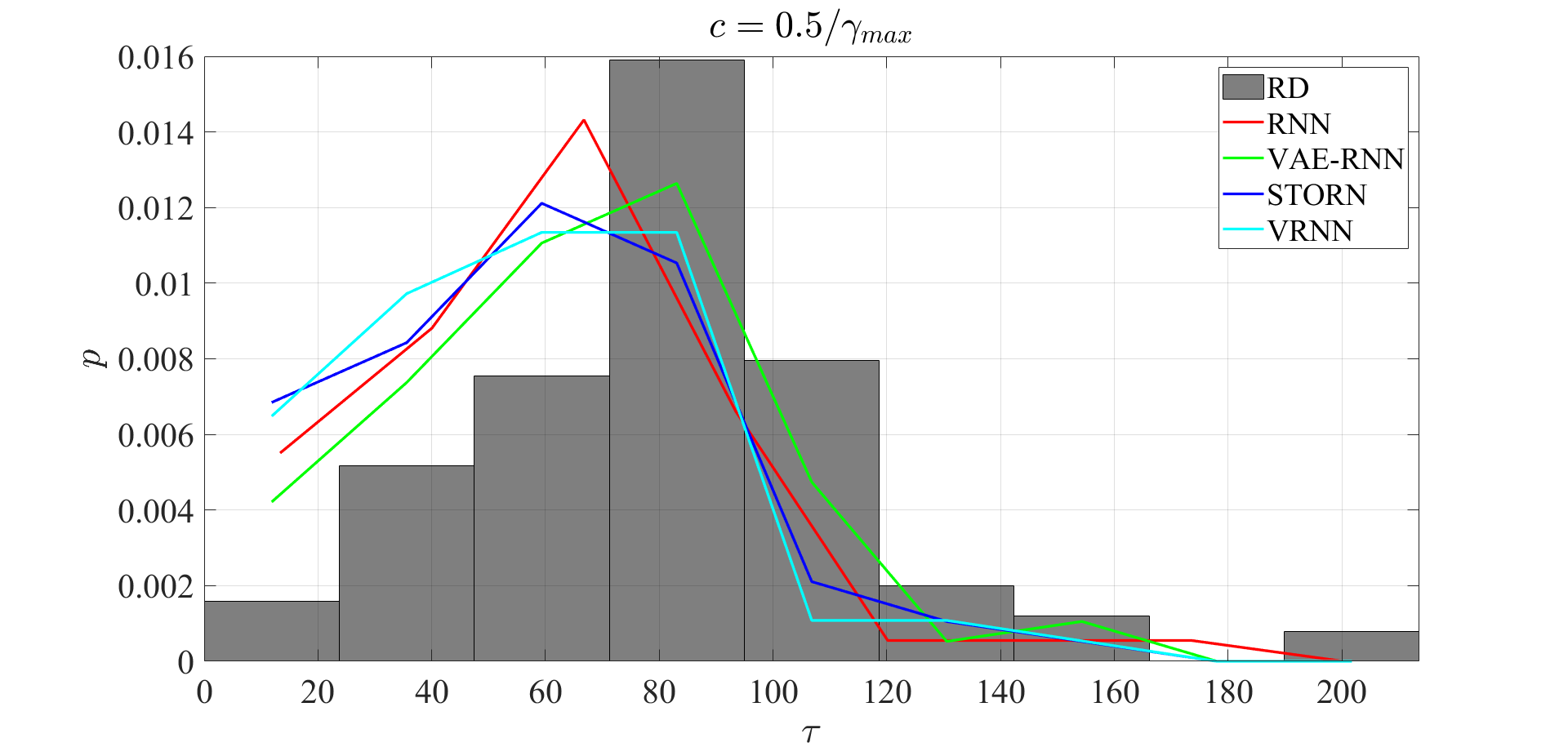} 
           \includegraphics[width=0.45\textwidth]{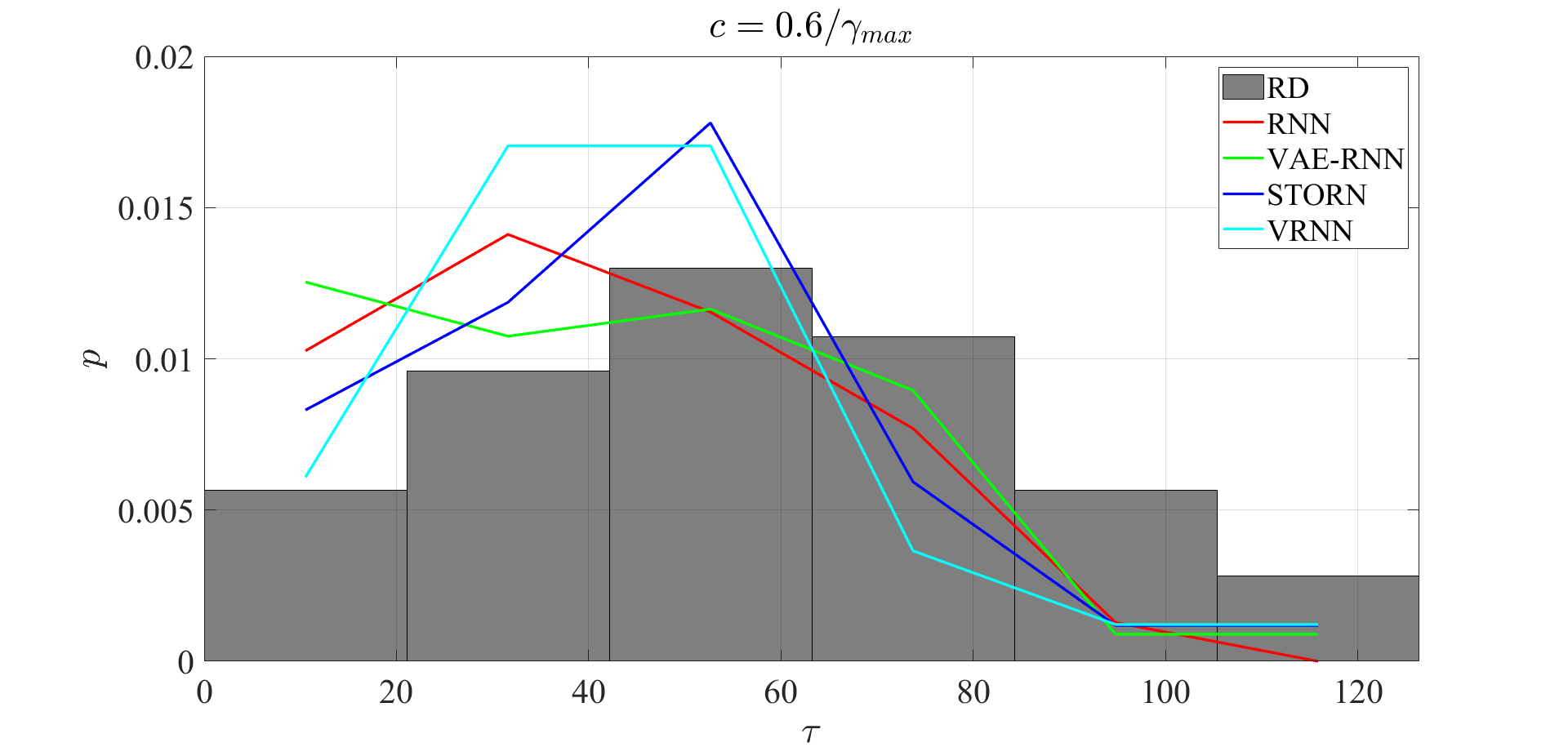} \\
              ~\ ~\  ~\ ~\ ~\ ~\ ~\ ~\ ~\ ~\ ~\ ~\ ~\ ~\ ~\ (f)  ~\  ~\ ~\ ~\ ~\ ~\ ~\ ~\ ~\ ~\ ~\ ~\ ~\ ~\  ~\ ~\ ~\ ~\  ~\ ~\ ~\ ~\ ~\ ~\ ~\ ~\  ~\ ~\ ~\ ~\ (g) \\
    \end{tabular}
   
    \caption{Zonally averaged stream function $\bar{\psi}_1$ (a). Frequency, expected duration, and variance of high amplitude excursions of $\gamma_1$ as a function of threshold $c$ (b). Probability density function of $\tau$ for fixed values of $c$ (c-g).}
    \label{fig:gamma_stats_psi1}
\end{figure}

\FloatBarrier
\newpage

\bibliography{references_no_url,leo_references}
\end{document}